\documentclass[runningheads]{llncs}

\usepackage[T1]{fontenc}
\usepackage{graphicx}
\usepackage{booktabs}            
\usepackage{subcaption}          
\usepackage{bm}                  
\usepackage[dvipsnames]{xcolor}  
\usepackage[fixed]{fontawesome5} 
\usepackage{amsfonts}            
\usepackage{amsmath}             
\usepackage{wrapfig}             
\usepackage{multirow}
\usepackage{float}
\usepackage{comment}
\usepackage[colorlinks=true, urlcolor=cyan]{hyperref}
\usepackage{caption} 
\usepackage{tikz}
\usetikzlibrary{shapes, arrows, arrows.meta, bayesnet, chains, backgrounds, fit, positioning}
\usetikzlibrary[calc]

\newcommand{\mainColor}{Dandelion}

\newcommand{\fsclvr}{\texttt{FS-CLVR}}
\newcommand{\fsclvrroom}{\texttt{FS-CLVR-room}}
\newcommand{\fsclvrdark}{\texttt{FS-CLVR-dark}}
\newcommand{\ycbood}{\texttt{YCB-OOD}}

\newcommand{\Shift}{\mathbf{x}}
\newcommand{\Angle}{\theta}
\newcommand{\Scale}{\mathbf{s}}
\newcommand{\Pattern}{\mathbf{p}}
\newcommand{\Color}{\mathbf{c}}
\newcommand{\Ambient}{k^a}
\newcommand{\Diffuse}{k^d}
\newcommand{\Specular}{k^s}
\newcommand{\Shininess}{\alpha}
\newcommand{\Classes}{\kappa}

\newcommand{\Scene}{\psi}
\newcommand{\Render}{\mathcal{I}_{\mathcal{R}}}
\newcommand{\Renderer}{\mathcal{R}}
\newcommand{\ImageData}{\mathcal{I}_{\mathcal{D}}}
\newcommand{\ImagePairSet}{P}
\newcommand{\FeatureChannel}{M}
\newcommand{\OptimalFeatureChannelSet}{M}
\newcommand{\FeaturemapSize}{N}

\newcommand{\LightIntensity}{\mathbf{l}}
\newcommand{\Floor}{F}
\newcommand{\Global}{\mathcal{G}}
\newcommand{\Object}{\mathcal{O}}
\newcommand{\Loss}{\mathcal{L}}

\newcommand{\Featuremap}{\mathrm{F}}
\DeclareMathOperator*{\argmin}{arg\,min} 

\newcommand{\Likelihood}{\mathrm{L}}
\newcommand{\NeuralLikelihood}{\mathrm{L}_{\mathrm{N}}}
\newcommand{\ColorLikelihood}{\mathrm{L}_{\mathrm{C}}}

\newcommand{\Observation}{\mathcal{I}}
\newcommand{\GenerativeObject}{\mathcal{O}}
\newcommand{\ScalarColor}{c}
\newcommand{\GenerativeMaterial}{\mathcal{M}}
\newcommand{\GenerativeTransform}{\mathcal{A}}

\newcommand{\GenerativeShape}{\mathcal{S}}

\newcommand{\LogNormalDistribution}{\mathrm{Log}\mathcal{N}}
\newcommand{\TruncatedNormalDistribution}{\mathrm{Tr}\mathcal{N}}

\newcommand{\Material}{\mathcal{M}}
\newcommand{\DiffGraphics}{\mathcal{R}}
\newcommand{\ProbabilisticParameters}{\Omega}
\newcommand{\ProtoVariables}{\psi_{\mathcal{S}}}
\newcommand{\merge}{\texttt{merge}}

\newcommand{\distance}{\texttt{distance}}

\begin{document}
\title{Bayesian Inverse Graphics \\ for Few-Shot Concept Learning}

\author{Octavio Arriaga \inst{1} \and Jichen Guo \inst{1} \and Rebecca Adam \inst{3} \and Sebastian Houben \inst{2} \and Frank Kirchner \inst{1,3}}
\authorrunning{Octavio Arriaga et al.}
\institute{Robotics Research Group, University of Bremen \\ \email{arriagac@uni-bremen.de \faIcon{envelope}}
\and University of Applied Sciences, Bonn-Rhein-Sieg 
\and
Robotics Innovation Center, DFKI GmbH
}
\maketitle

\begin{abstract}
    Humans excel at building generalizations of new concepts from just one single example. 
    Contrary to this, current computer vision models typically require large amount of training samples to achieve a comparable accuracy.
    In this work we present a Bayesian model of perception that learns using only minimal data, a prototypical probabilistic program of an object.
    Specifically, we propose a generative inverse graphics model of primitive shapes, to infer posterior distributions over physically consistent parameters from one or several images.
    We show how this representation can be used for downstream tasks such as few-shot classification and pose estimation.
    Our model outperforms existing few-shot neural-only classification algorithms and demonstrates generalization across varying lighting conditions, backgrounds, and out-of-distribution shapes. 
    By design, our model is uncertainty-aware and uses our new differentiable renderer for optimizing global scene parameters through gradient descent, sampling posterior distributions over object parameters with Markov Chain Monte Carlo (MCMC), and using a neural based likelihood function
    \protect\footnotemark[1].
\end{abstract}

\footnotetext[1]{Datasets and code available at
    \href{https://github.com/oarriaga/bayesian-inverse-graphics}{github.com/oarriaga/bayesian-inverse-graphics}
}

\section{Introduction}~\label{sec:introduction}
    Children have the remarkable ability to learn new concepts from only a small set of examples~\cite{gopnik2012scientific,xu2007word,lake2011one}.
Replicating this human capacity has been a long standing challenge within the few-shot learning research community, and has been considered a milestone for building machines capable of having the same flexibility and learning capacity of humans~\cite{tenenbaum2011grow,lake2017building}.
Current deep learning models hold state of the art results in many few-shot learning tasks, owning great part of their success to the unprecedented availability of large datasets and computational resources~\cite{lecun2015deep,hooker2021hardware}.
This has resulted in large language models (LLMs) and vision transformers (ViT)~\cite{dosovitskiy2020image} showing realistic generative capabilities~\cite{ramesh2021zero}, as well as zero-shot task generalization.
These new deep learning paradigm contains architectures with billions of parameters, which are optimized over billions of data samples.
For instance, a generic vision model like SAM~\cite{kirillov2023segment} contains more than half a billion parameters, which were optimized using more than one billion segmentation masks.

Despite the striking results of these models, this high sample complexity still remains exceptionally large when compared to the learning ability of humans~\cite{lake2017building}.
Few-shot learning methods aim to reduce this sample and model complexity by extending learning algorithms with meta-learning, composition and intuitive physics~\cite{lake2015human}. 
However, DL models rarely use composable structures that are physically consistent.
Rather, they are often justified by meta design choices, such as optimization ease through residual connections~\cite{he2016deep}, or prevention of feature information loss through densely connected layers~\cite{huang2017densely} or multiple featuremap resolutions~\cite{wang2020deep,ronneberger2015u}. 
Moreover, these design choices are often validated only through predictive accuracy which disregards any form of uncertainty quantification, eventually leading into uncalibrated predictions~\cite{valdenegro2021find}.
Although these issues are discussed within some of the few-shot learning literature~\cite{lake2015human,lake2019omniglot}, few-shot learning datasets continue to favour large parametric models by having large training datasets; thus, defying the purpose of learning from only few data points.
\begin{figure}[t!]
    \centering
    \includegraphics[width=1.0\linewidth]{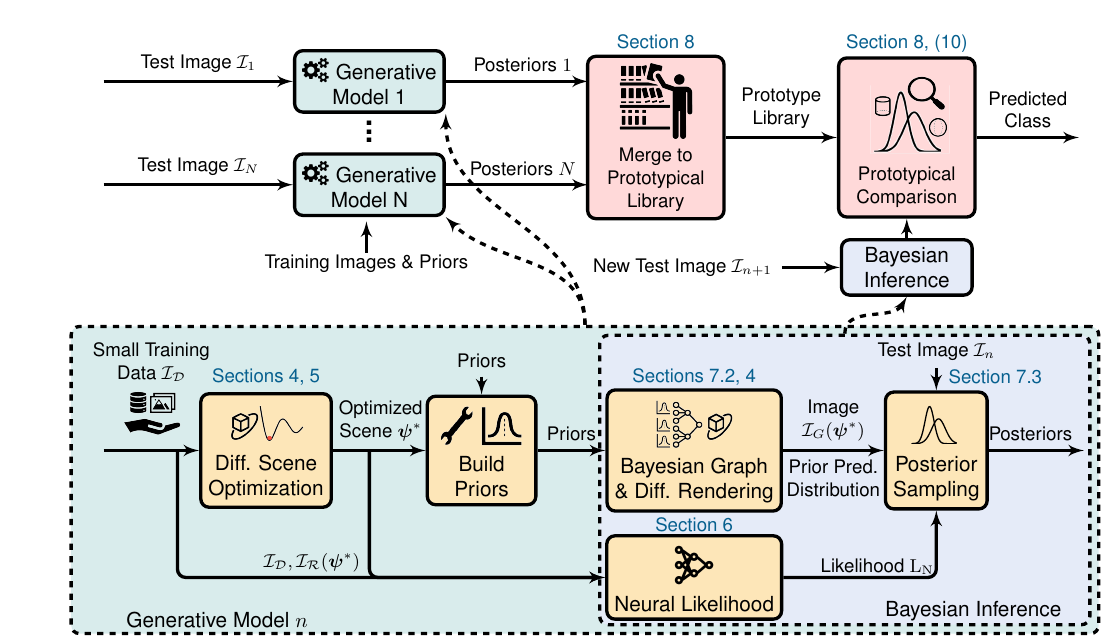}
    \caption{Neuro-symbolic inverse graphics model for few-shot learning}~\label{fig:global_model_architecture}
    \vspace{-0.80cm}
\end{figure}

In order to address the open challenges within the current paradigm of computing large point estimates using billions of samples, we propose exploring an opposite question: How can we build the smallest uncertainty aware vision model that can generalize from only few training images?
We approach this question using an inverse graphics framework based on probabilistic cognitive models~\cite{probmods2,rule2020child,ghahramani2015probabilistic,griffiths2010probabilistic}.
Specifically, we apply the \textit{Bayesian workflow}~\cite{gelman2020bayesian} to build a probabilistic generative model (PGM) to simulate the data generative process of images~\cite{kulkarni2015deep}.
We test our model through prior and posterior predictive checks, which aim to mitigate any possible bias, or incorrect model assumption~\cite{martin2022bayesian}.
By explicitly using a physics-based model, we reduce the model complexity to a short parametric description that allows us to estimate the full posterior of our parameters.
Figure~\ref{fig:global_model_architecture} shows an overview of our neuro-symbolic architecture for few-shot learning with minimal data.
Our main contributions include: 
\begin{itemize}
    \item[$\bullet$] A probabilistic generative model that allows us to infer distributions over the properties and poses of new unseen objects from a single image.
    \item[$\bullet$] A few-shot classification algorithm that uses posterior distributions to build prototypical probabilistic programs with an embedded distance function.
    \item[$\bullet$] A set of benchmarks that test generalization in a low training sample regime, under different lighting conditions, backgrounds, and novel unseen objects.
    \item[$\bullet$] A new differentiable renderer that is compatible with probabilistic programming languages, deep learning models and optimization libraries.
    \item[$\bullet$] Finally, the application of a probabilistic physics-based model does not remove the possibility of using data-driven methods; rather, we show that one can combine neural architectures with probabilistic symbolic representations through a neural color likelihood function, which outperforms each of these elements separately.
\end{itemize}

\section{Related Work}\label{sec:related_work}
    \begin{table}[b!]
    \centering
    \caption{Few-shot classification datasets with their number of samples}~\label{table:few_shot_datasets}
    \begin{tabular}{lr|lr}
        \toprule
        {Classification}      & {samples} & {Pose estimation}    & {samples} \\
        \midrule
        mini-imagenet~\cite{vinyals2016matching}   & 60K    &  Latent fusion~\cite{park2020latentfusion} & 480M \\ 
        CIFAR-FS~\cite{bertinetto2018meta}         & 60K    &  OnePose~\cite{sun2022onepose}             & 128K \\ 
        FGVCAircraft~\cite{maji13fine-grained}     & 10K    &  ShapeNet6D~\cite{he2022fs6d}              & 800K \\
        FGVCFungi~\cite{fungi-challenge-fgvc-2018} & 89K    &  \textbf{CLEVR-FS (ours)}                  & 300  \\
        Omniglot~\cite{lake2015human}              & 32K    &  \textbf{CLEVR-FS dark (ours)}             & 300  \\
        Omniglot small~\cite{lake2015human}        & 2.7K   &  \textbf{CLEVR-FS room (ours)}             & 300  \\
        \bottomrule
    \end{tabular}
\end{table}
Few-shot learning models have been classified into metric learning, meta-learning, memory-augmented networks and generative models~\cite{wang2020generalizing}.
One of the most relevant models under the metric learning classification is the prototypical network model~\cite{snell2017prototypical}.
This model is trained using meta learning episodes that build new classification problems at every optimization step.
The model learns to embed images into a latent vector space that is reused to classify new samples based on their distance to the projected mean of the support classes.
Other meta-learning neural algorithms perform a double optimization loop that updates the model weights within episodes and training samples~\cite{finn2017model,nichol2018first}.
The method most similar to our approach is the influential Bayesian program learning (BPL) model~\cite{lake2015human}.
This method builds a meta generative model of characters by building stochastic programs that are optimized under a Bayesian criterion. 
In a revision of the state-of-the-art few-shot algorithms, the BPL model still outperformed neural variants in the Omniglot dataset~\cite{lake2019omniglot}.
However, the original BPL algorithm was created as a generative meta-program of characters, and is not directly applicable to 3D geometric objects under realistic lighting conditions.
Analogously to few-shot classification algorithms, neural few-shot pose estimation methods also rely to a greater extend in large training datasets.
Moreover, in contrast to the few-shot classification task, there is no clear benchmark dataset for few-shot pose estimation.
As a consequence, they are often tested on different datasets precluding a fair evaluation.
Some of the most relevant models are FS6D~\cite{he2022fs6d}, OnePose~\cite{sun2022onepose} and LatentFusion~\cite{park2020latentfusion}.
The corresponding number of images used for training these few-shot learning methods are listed in Table~\ref{table:few_shot_datasets}.

\section{Minimal-data Benchmarks}~\label{sec:few_shot_datasets}
\begin{figure}[t!]
    \begin{subfigure}[T]{0.49\textwidth}
        \caption*{20-way 1-shot learning}
        \begin{subfigure}[b]{\textwidth}
            \centering
            \begin{subfigure}[b]{0.185\textwidth}
                \includegraphics[width=\textwidth]{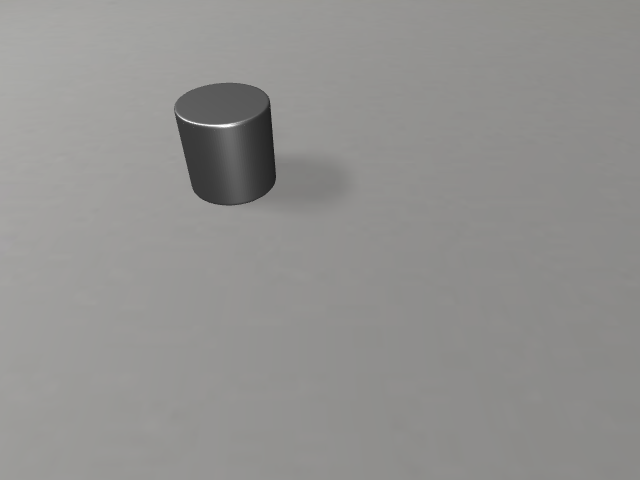}
            \end{subfigure}
            \begin{subfigure}[b]{0.185\textwidth}
                \includegraphics[width=\textwidth]{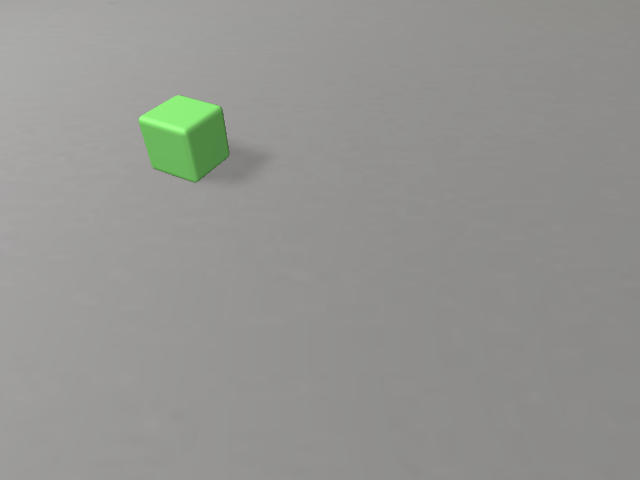}
            \end{subfigure}
            \begin{subfigure}[b]{0.185\textwidth}
                \includegraphics[width=\textwidth]{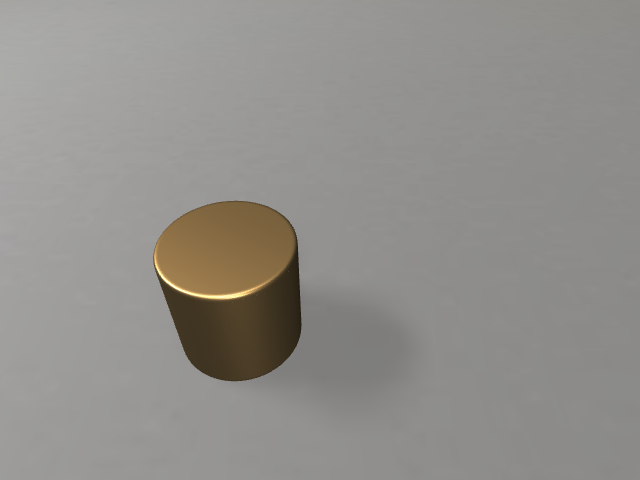}
            \end{subfigure}
            \begin{subfigure}[b]{0.185\textwidth}
                \includegraphics[width=\textwidth]{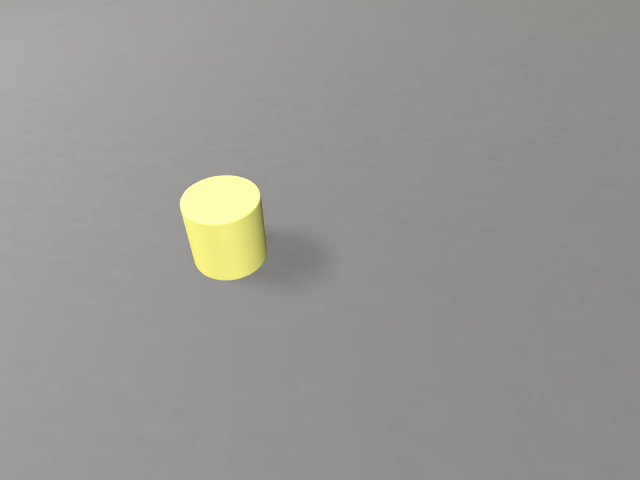}
            \end{subfigure}
            \begin{subfigure}[b]{0.185\textwidth} 
                \includegraphics[width=\textwidth]{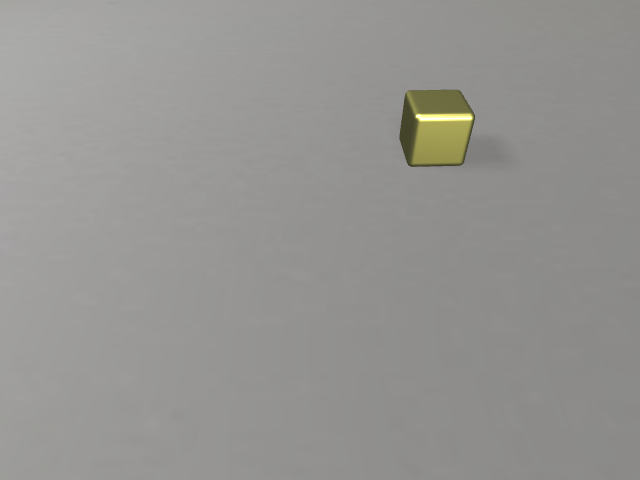}
            \end{subfigure}

            \begin{subfigure}[b]{0.185\textwidth} 
                \includegraphics[width=\textwidth]{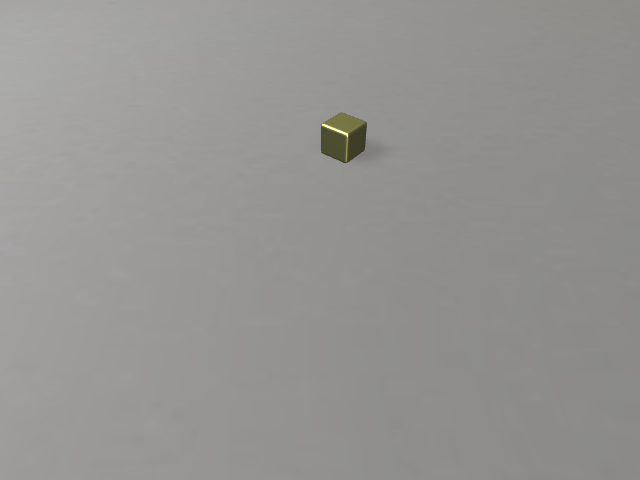}
            \end{subfigure}
            \begin{subfigure}[b]{0.185\textwidth}
                \includegraphics[width=\textwidth]{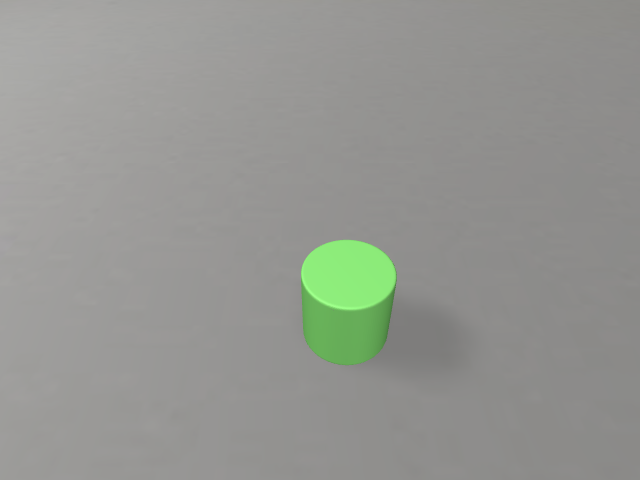}
            \end{subfigure}
            \begin{subfigure}[b]{0.185\textwidth}
                \includegraphics[width=\textwidth]{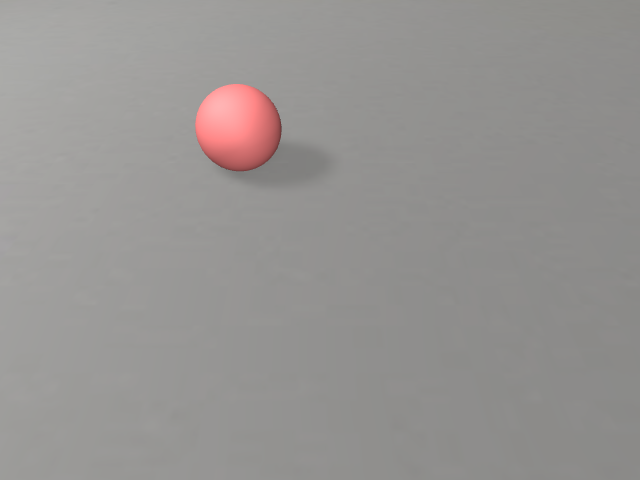}
            \end{subfigure}
            \begin{subfigure}[b]{0.185\textwidth}
                \includegraphics[width=\textwidth]{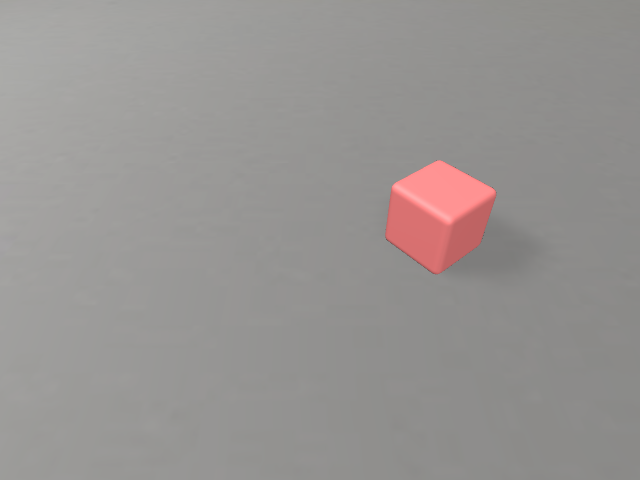}
            \end{subfigure}
            \begin{subfigure}[b]{0.185\textwidth}
                \includegraphics[width=\textwidth]{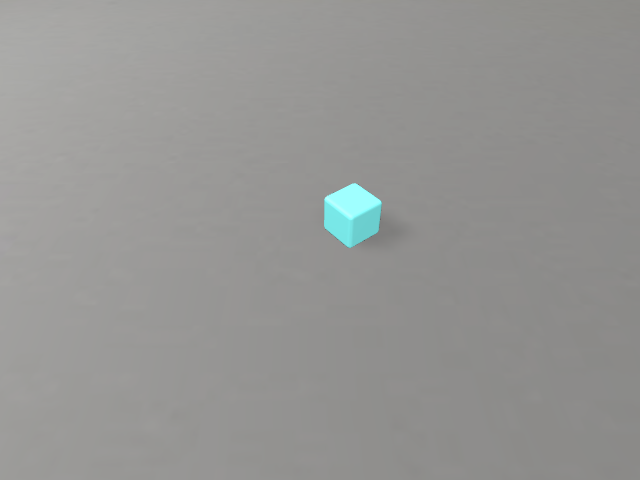}
            \end{subfigure}

            \begin{subfigure}[b]{0.185\textwidth}
                \includegraphics[width=\textwidth]{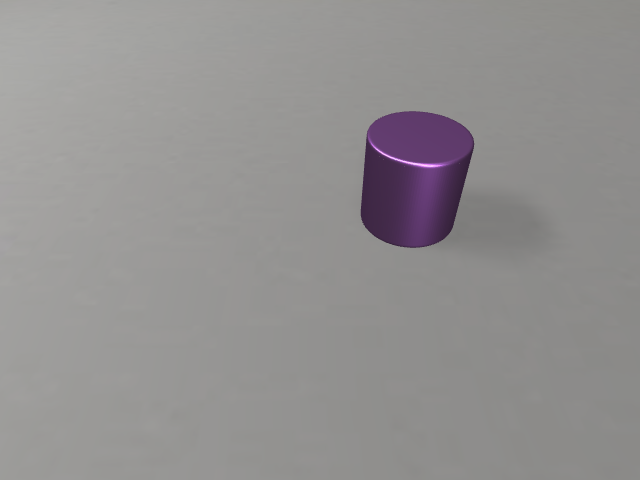}
            \end{subfigure}
            \begin{subfigure}[b]{0.185\textwidth}
                \includegraphics[width=\textwidth]{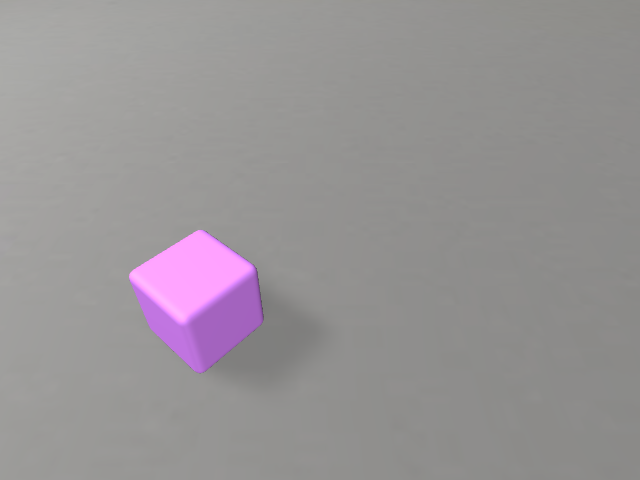}
            \end{subfigure}
            \begin{subfigure}[b]{0.185\textwidth}
                \includegraphics[width=\textwidth]{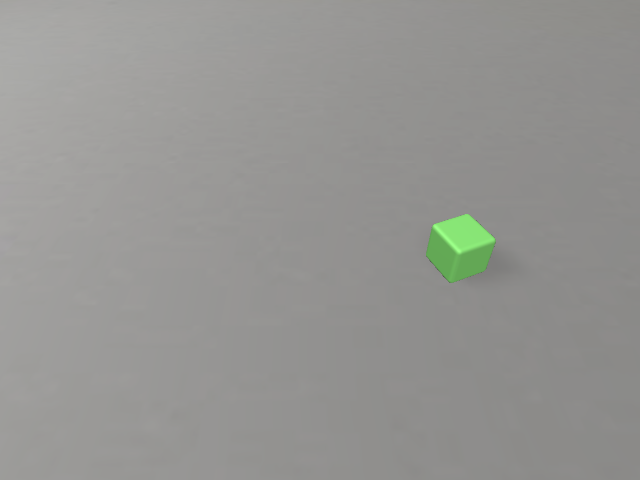}
            \end{subfigure}
            \begin{subfigure}[b]{0.185\textwidth}
                \includegraphics[width=\textwidth]{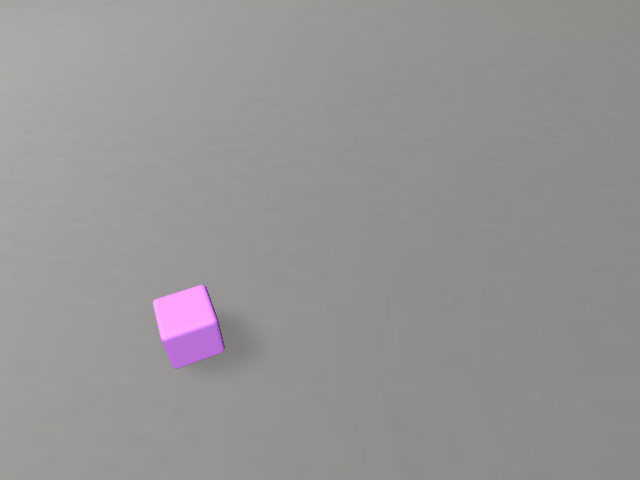}
            \end{subfigure}
            \begin{subfigure}[b]{0.185\textwidth}
                \includegraphics[width=\textwidth]{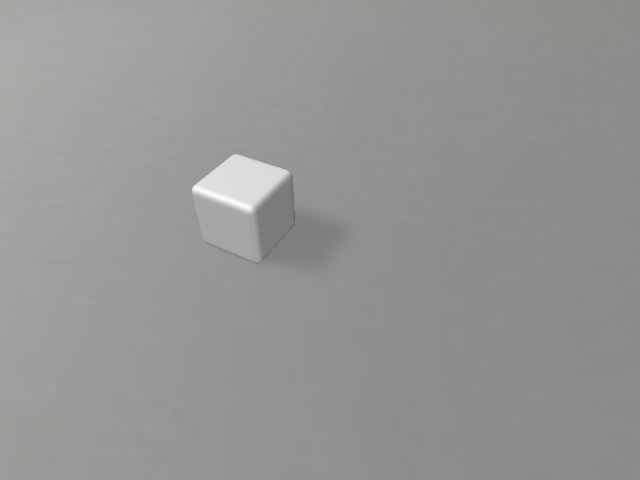}
            \end{subfigure}

            \begin{subfigure}[b]{0.185\textwidth}
                \includegraphics[width=\textwidth]{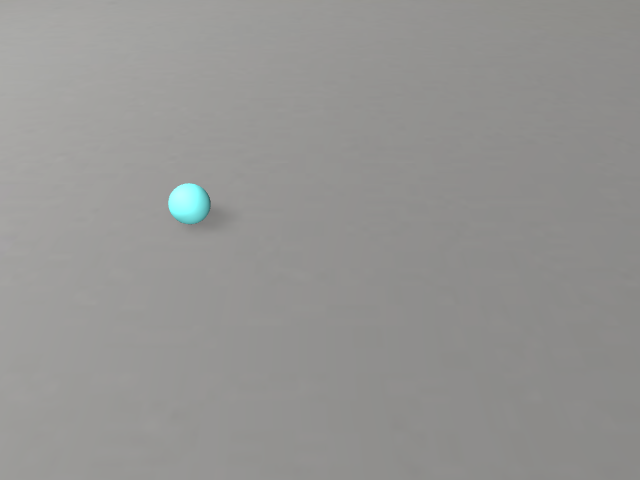}
            \end{subfigure}
            \begin{subfigure}[b]{0.185\textwidth}
                \includegraphics[width=\textwidth]{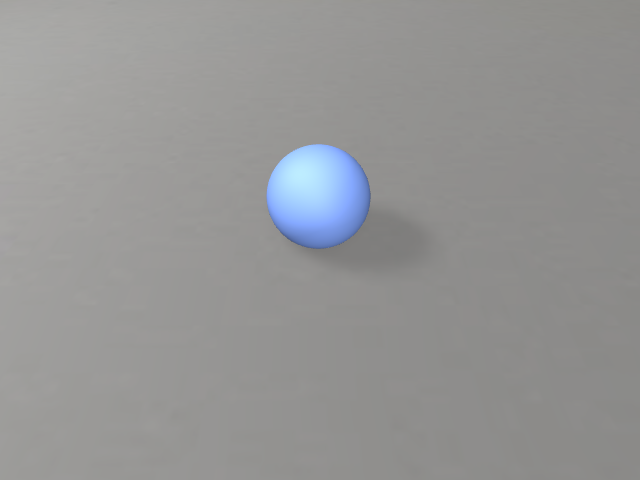}
            \end{subfigure}
            \begin{subfigure}[b]{0.185\textwidth}
                \includegraphics[width=\textwidth]{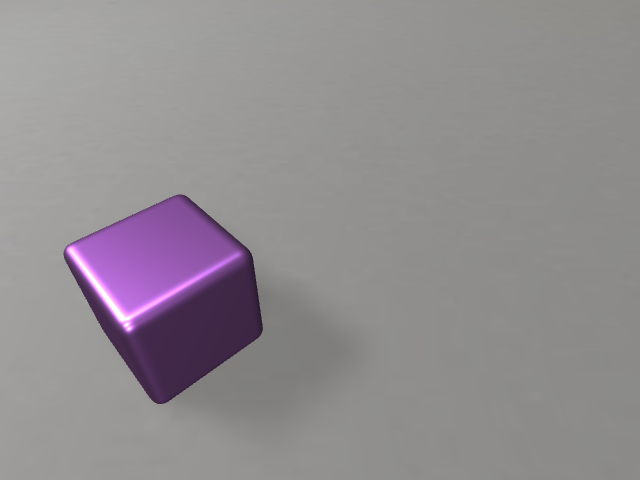}
            \end{subfigure}
            \begin{subfigure}[b]{0.185\textwidth}
                \includegraphics[width=\textwidth]{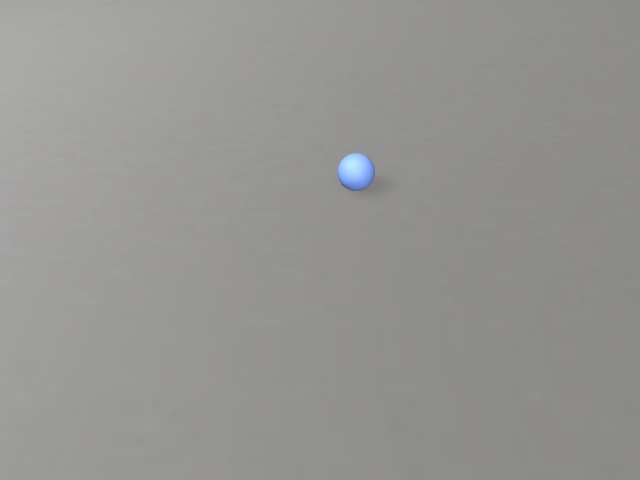}
            \end{subfigure}
            \begin{subfigure}[b]{0.185\textwidth}
                \includegraphics[width=\textwidth]{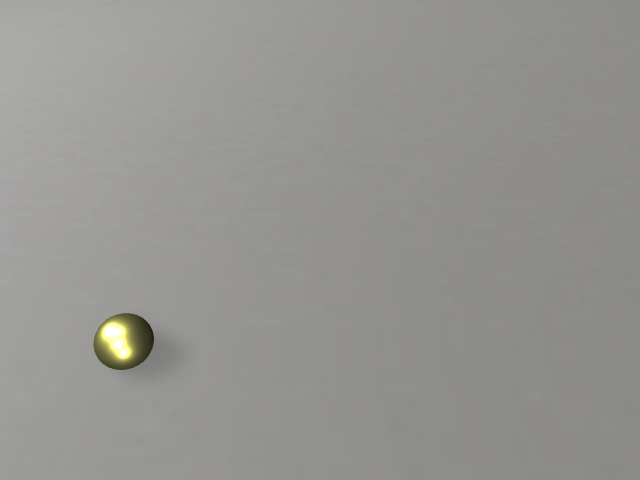}
            \end{subfigure}
        \end{subfigure}
            \centering
            \caption*{To which class does this belong to?\protect\footnotemark[2]}
            \begin{subfigure}[b]{0.185\textwidth}
                \includegraphics[width=\textwidth]{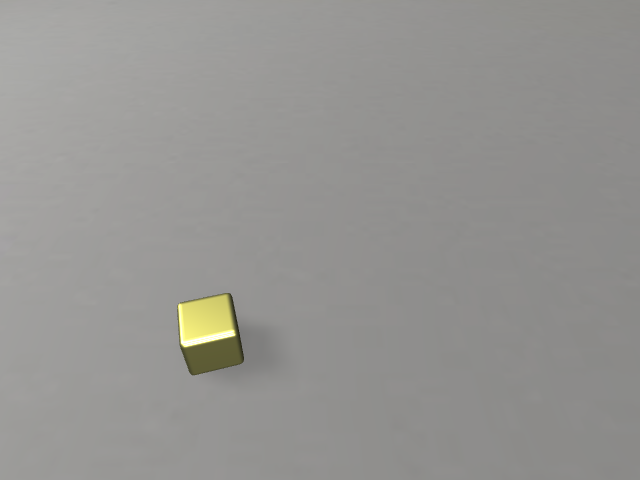}
            \end{subfigure}
            \caption{Few-shot training episode}~\label{fig:few_shot_episode}
    \end{subfigure}
    ~
    \begin{subfigure}[T]{0.49\textwidth}
        \centering
        \begin{subfigure}[b]{0.235\textwidth}
            \caption*{\normalsize{\fsclvr}}
            \includegraphics[width=\textwidth]{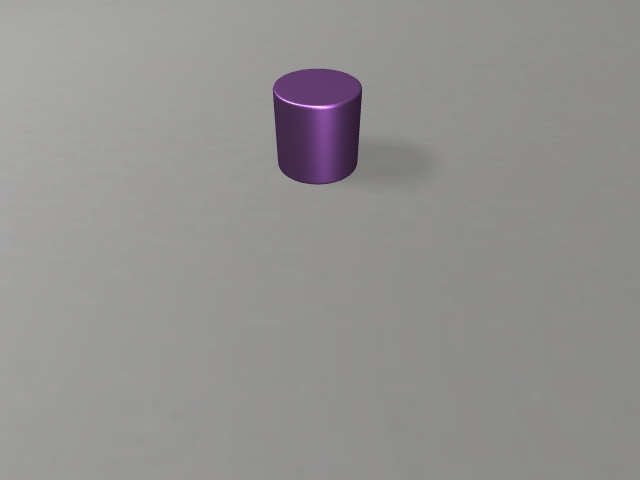}
        \end{subfigure}
        \begin{subfigure}[b]{0.235\textwidth}
            \caption*{\normalsize{\texttt{DARK}}}
            \includegraphics[width=\textwidth]{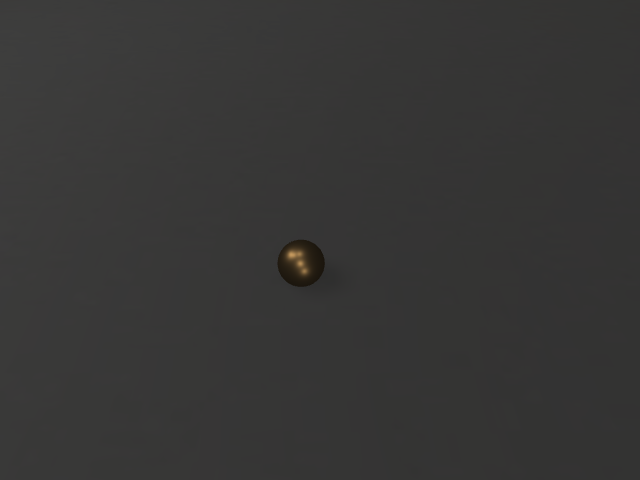}
        \end{subfigure}
        \begin{subfigure}[b]{0.235\textwidth}
            \caption*{\normalsize{\texttt{ROOM}}}
            \includegraphics[width=\textwidth]{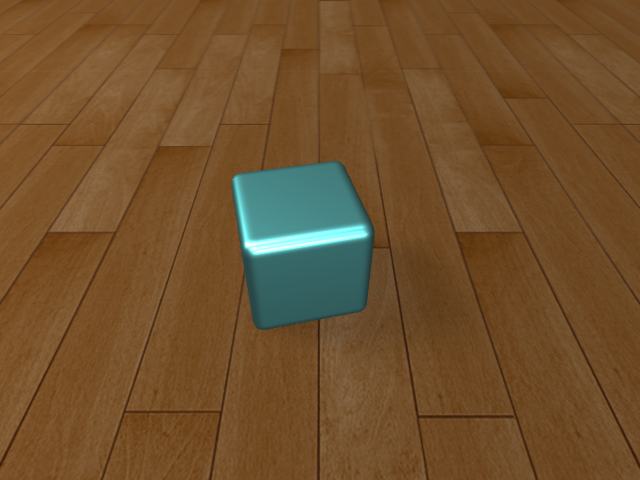}
        \end{subfigure}
        \begin{subfigure}[b]{0.235\textwidth}
            \caption*{\normalsize{\ycbood}}
            \includegraphics[width=\textwidth]{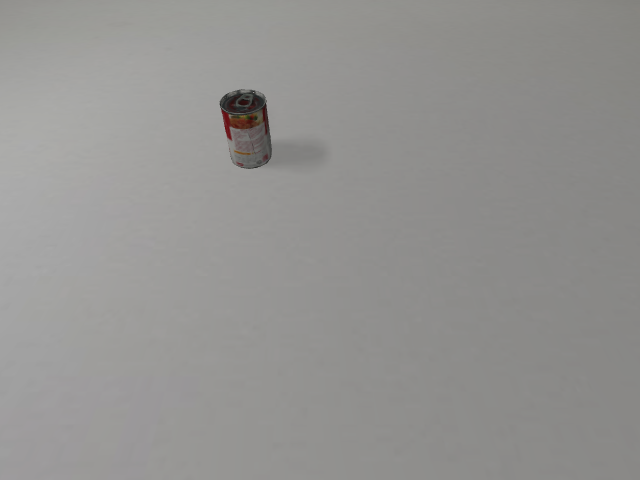}
        \end{subfigure}

        \begin{subfigure}[b]{0.235\textwidth}
            \includegraphics[width=\textwidth]{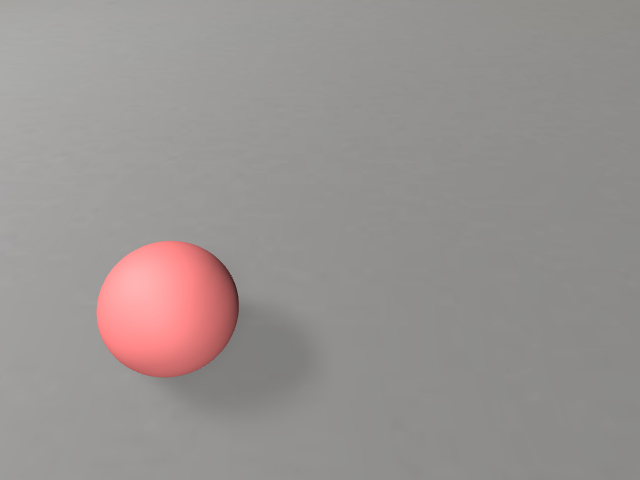}
        \end{subfigure}
        \begin{subfigure}[b]{0.235\textwidth}
            \includegraphics[width=\textwidth]{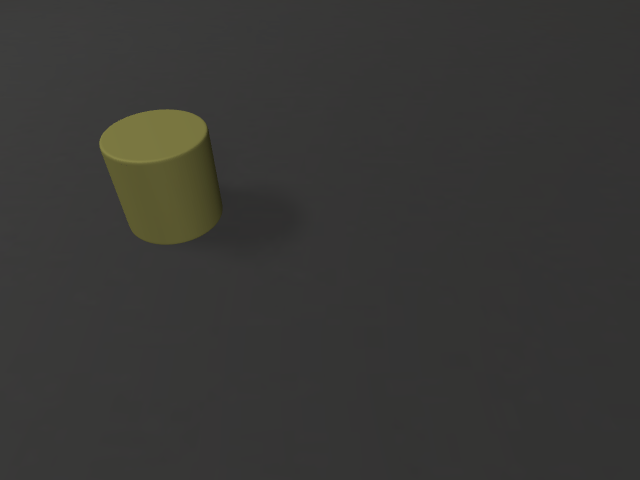}
        \end{subfigure}
        \begin{subfigure}[b]{0.235\textwidth}
            \includegraphics[width=\textwidth]{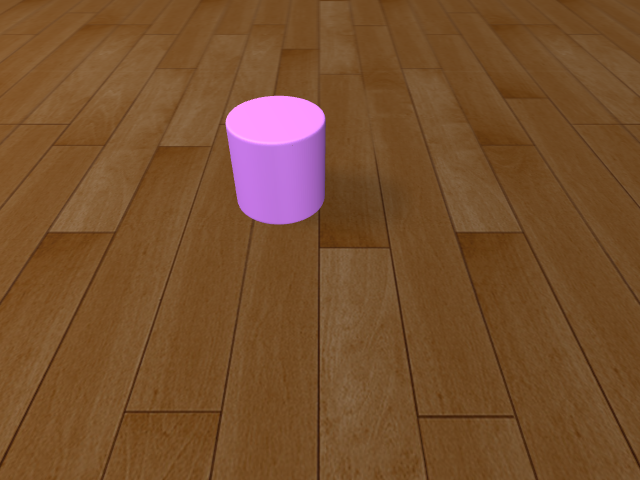}
        \end{subfigure}
        \begin{subfigure}[b]{0.235\textwidth}
            \includegraphics[width=\textwidth]{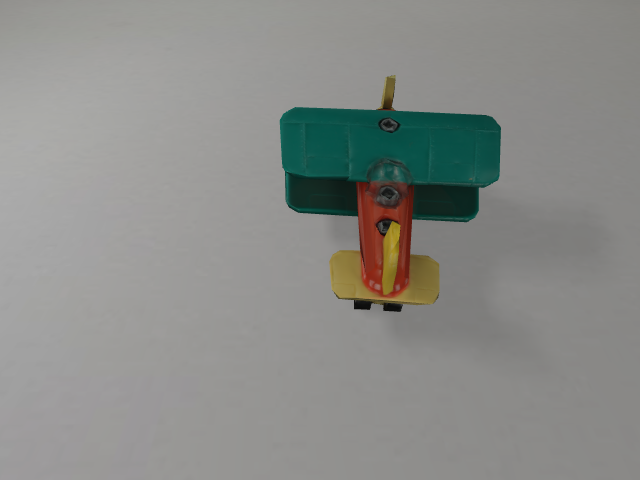}
        \end{subfigure}

        \begin{subfigure}[b]{0.235\textwidth}
            \includegraphics[width=\textwidth]{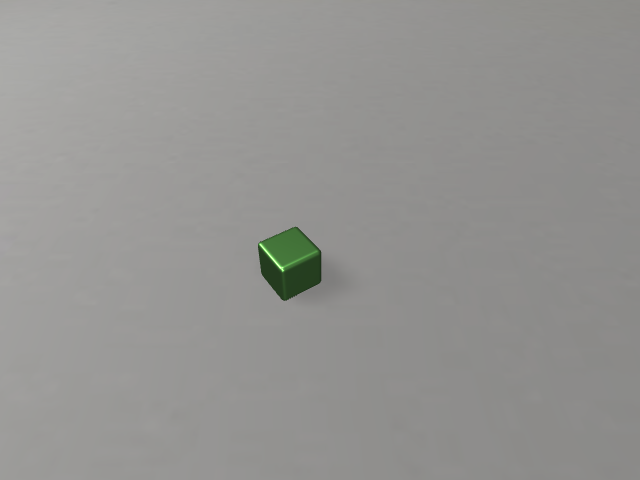}
        \end{subfigure}
        \begin{subfigure}[b]{0.235\textwidth}
            \includegraphics[width=\textwidth]{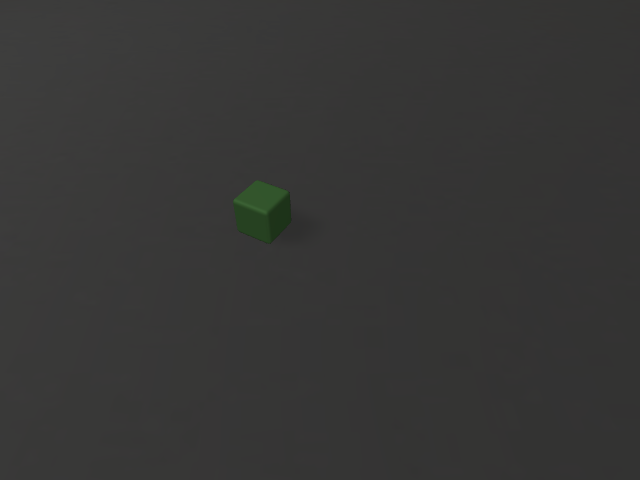}
        \end{subfigure}
        \begin{subfigure}[b]{0.235\textwidth}
            \includegraphics[width=\textwidth]{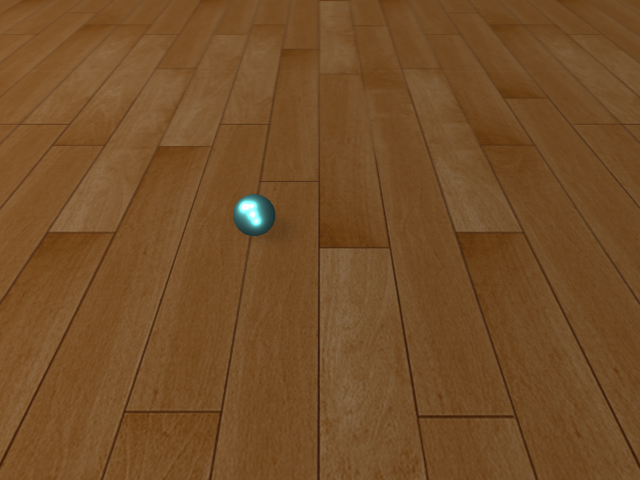}
        \end{subfigure}
        \begin{subfigure}[b]{0.235\textwidth}
            \includegraphics[width=\textwidth]{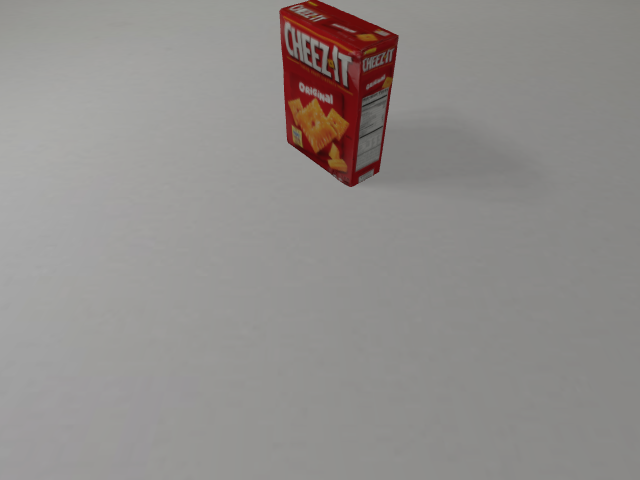}
        \end{subfigure}

        \begin{subfigure}[b]{0.235\textwidth}
            \includegraphics[width=\textwidth]{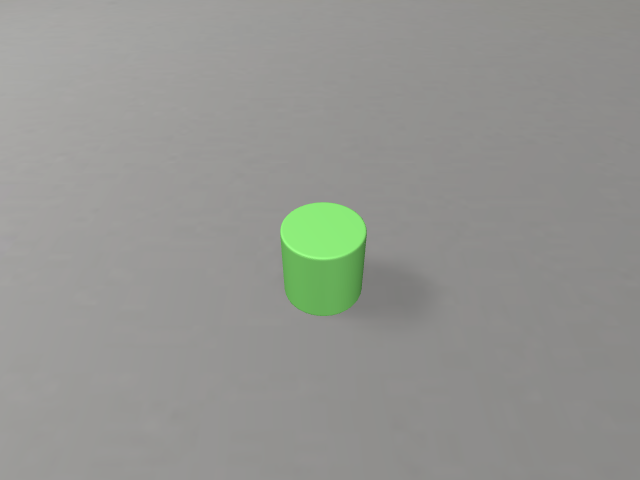}
        \end{subfigure}
        \begin{subfigure}[b]{0.235\textwidth}
            \includegraphics[width=\textwidth]{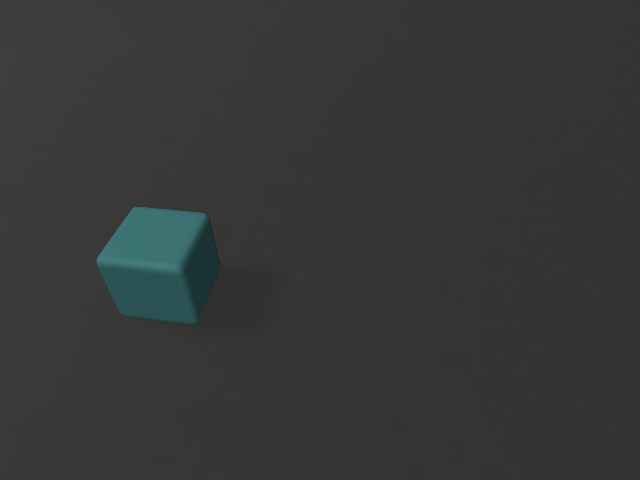}
        \end{subfigure}
        \begin{subfigure}[b]{0.235\textwidth}
            \includegraphics[width=\textwidth]{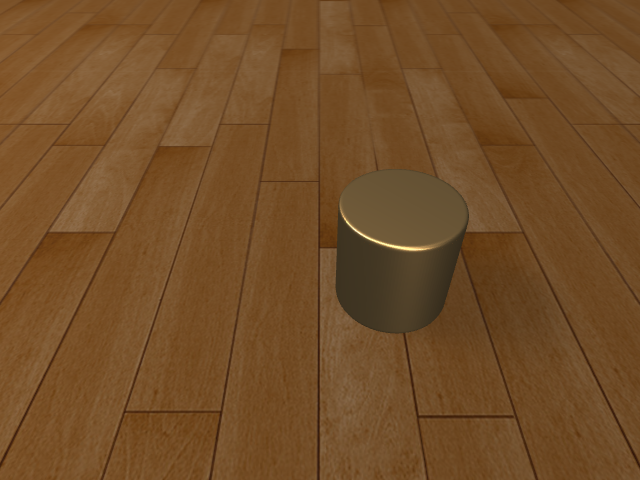}
        \end{subfigure}
        \begin{subfigure}[b]{0.235\textwidth}
            \includegraphics[width=\textwidth]{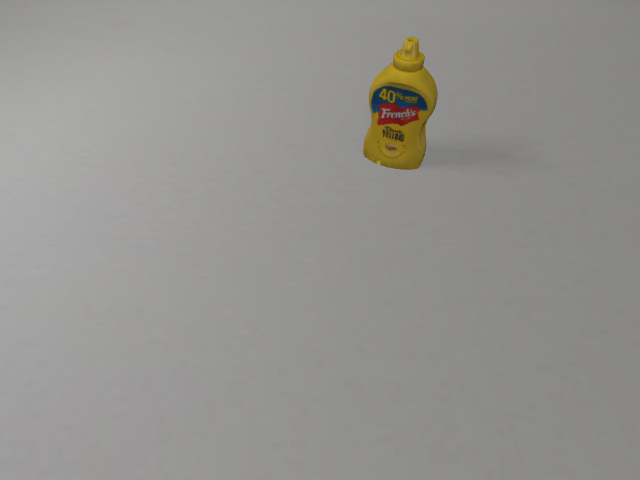}
        \end{subfigure}
        \caption*{Minimal data benchmarks}~\label{fig:primitives}
    \end{subfigure}
    \caption{Samples of our few-shot training datasets}
    \label{fig:few_shot_example_and_datasets}
    \vspace{-0.25cm}
\end{figure}

    As indicated in Table~\ref{table:few_shot_datasets}, current few-shot datasets contain a large amount of training samples; thus, undermining the ability of current few-shot models to learn from only few data points.
Under this consideration, we adapted the ubiquitously employed CLEVR dataset~\cite{johnson2017clevr} for few-shot learning and few-shot pose estimation.
Specifically, we present the following 4 benchmarks \fsclvr, \fsclvrroom, \fsclvrdark~and \ycbood.
These benchmarks assess in a controllable environment the generalization of few-shot learning models when using only minimal data.
All samples include their respective 6D poses making it a suitable benchmark for few-shot pose estimation models.
The \fsclvrroom, \fsclvrdark~and \ycbood~validate respectively the model generalization to new backgrounds, darker lighting conditions, and to out-of-distribution (OOD) complex shapes.
These \texttt{FS-CLVR} datasets were rendered using the same shape, materials and colors employed in CLEVR, as well as a vertical field of view (VFOV) of $42.5^{\circ}$ present in most commercial depth cameras.
Each of these benchmarks contain 300 images separated into 30 training classes and 20 test classes, each class having 6 shot images.
The~\ycbood~has no training samples and it contains 10 test classes with 6 shots each.
This dataset is meant to validate the OOD generalization of few-shot models to unseen classes, and consists of the following 10 classes from the YCB dataset~\cite{calli2015benchmarking}: power-drill, tomato-soup, airplane-A, foam brick, softball, apple, cracker box, mustard bottle, tuna fish can and mug. 
Furthermore, while these datasets mostly consist of primitive shapes, they still remain challenging to generic perception algorithms and deep learning models.
Specifically, they pose the following open problems: finite and infinite symmetries, textureless objects, reflective materials, changing lighting conditions and ultimately few number of training samples.
Finally, as shown in Figure~\ref{fig:few_shot_episode} these few shot tasks can remain challenging for humans.

\section{Merging Bayesian Inference, Graphics and DL}\label{sec:differentiable_rendering}
Our inverse graphics model uses elements from three different fields: Bayesian inference, computer graphics, and deep learning. 
This allows us to reduce the weaknesses of each model by using their complementary strengths.
For example, most DL architectures ignore the best known physical descriptions of the problems they aim to solve~\cite{lake2017building}.
Specifically, most computer vision tasks take as input a set of color images from different viewpoints and lighting conditions, and aim to extract information about the physical world.
However, most DL vision models don't use any physical simulation between light and matter~\cite{spielberg2023differentiable}. 
In contrast to DL, our model does incorporate this explicit physical knowledge by using a computer graphics pipeline that also encodes known physical limits as prior distributions.
Furthermore, many physical phenomena can be described using different models that compromise between accurate predictions and computational complexity. 
Fortunately, the computer graphics community has been developing the right abstractions, algorithms, and hardware, to efficiently simulate and optimize these models in a physically consistent manner~\cite{pharr2023physically,zhao2020physics,Mitsuba3}.
Our model uses a common physical approximation known as ray-tracing~\cite{Shirley2024RTW1,buck2019ray}, in which an image is rendered by simulating the intersection of light rays with the properties of a given scene, such as the geometry of the objects, or the location of different light sources. 
Finally, our renderer does not consider all nuances that real images could have.
Those can include complex material models (metallic-roughness), soft shadows, and smooth cornered shapes.
To address this gap between simulation and reality we propose a neural likelihood function that measures image similarity in features space.
\footnotetext[2]{2nd row, 1st column. The lack of additional samples (shots) prevents us to understand the relation between the object size and the scene's depth.} 

Moreover, we would like to emphasize that our approach is not limited to computer vision, and that the same framework can be applied to many machine learning problems.
This approach can be understood more broadly under the term \textit{Bayesian workflow}~\cite{martin2022bayesian,gelman2020bayesian}. 
In which a general recipe starts by collecting all the known information about the system of study, then encoding information and known physical limitations of the variables of such system through prior distributions. 
Then, validating all encoded elements by simulating forward the system and confirming that the outputted samples show what one expects to observe before any new evidence is presented. 
Subsequently, one performs Bayesian inference by conditioning the model on new observations.
Finally, one validates that the outputted posterior distributions correspond to a plausible solution.

\begin{figure}[t!]
    \begin{subfigure}[T]{0.5\linewidth}
        \includegraphics[width=\linewidth]{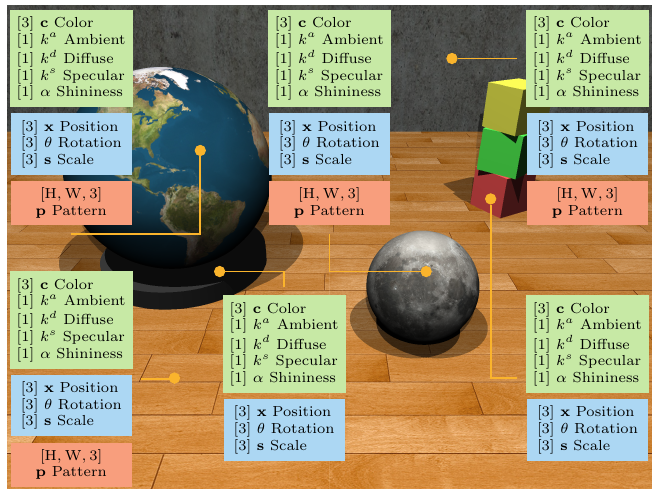}
        \caption{Differentiable scene parameters}~\label{fig:differentiable_rendering_parameters}
    \end{subfigure}
    \begin{subfigure}[T]{0.5\linewidth}
\begin{tikzpicture}[scale=0.75, transform shape, outer sep=0.01cm, node distance=0.8cm,]
        \tikzstyle{line} = [draw, -latex']
        \tikzstyle{parameter} = [minimum width=1cm, minimum height=0.7cm, align=center,
                                 rectangle, thick, draw=black!50, text width=1.9cm]
        \tikzstyle{function} = [minimum width=1cm, minimum height=0.7cm,
                                rectangle, thick, draw=black!50, text width=2.2cm, align=center]
        \tikzstyle{box} = [minimum width=1cm, minimum height=0.6cm,
                           rectangle, thick, draw=black!50]
        \tikzstyle{bigbox} = [draw=black!100, thick, rectangle]

        \tikzstyle{var} = [minimum width=1cm, minimum height=0.7cm, align=center, text width=1.9cm]

        \node[parameter] (light) {Lights \hfill \faLightbulb};
        \node[parameter, below of=light] (shape) {Shape \hfill \faShapes};
        \node[parameter, below of=shape] (pattern) {Pattern \hfill \faVolleyballBall};
        \node[parameter, below of=pattern] (material) {Material \hfill \faPencil*};

        \begin{pgfonlayer}{background}
            \node[bigbox] [fit = (light) (material)] (parameters) {};
        \end{pgfonlayer}

        \node[above= of parameters, yshift=-0.7cm] (title) {Scene parameters ($\Scene$)};

        \node[parameter, right= of parameters] (render) {$\Renderer(\Scene)$};
        \node[above= of render, yshift=-0.5cm] (function) {Render function};


        \node[parameter, right= of render] (loss) {$\Loss(\Render, \ImageData)$};
        \node[above= of loss, yshift=-0.5cm] (Loss) {Loss function};

        \node[var, above= of render, yshift=0.65cm] (data) {Observations ($\ImageData$)};

        \path [line] (parameters) -- (render);
        \path [line] (render) -- node[above,align=center]{$\Render$} (loss);
        \path [line] (data.east) -| (Loss.north);
        \path [line] (data.east) -| (Loss.north);

        \draw[line, dashed] ([yshift=-1.0cm]loss.south) -- ([yshift=-1.0cm]render.south);
        \node[below= of loss, yshift=0.5cm, xshift=-1.5cm] (autodiff) {
            Gradients $\nabla_{\Scene} \Loss (\Renderer(\Scene), \ImageData)$};
\end{tikzpicture}
        \vspace{0.8cm}
        \caption{Differentiable rendering pipeline}~\label{fig:differentiable_scene_optimization}
    \end{subfigure}
    \caption{Model based scene optimization}~\label{fig:model_based_optimization2}
    \vspace{-0.5cm}
\end{figure}

In this work we introduce a new differentiable ray tracer built entirely using JAX primitives~\cite{jax2018github}; thus, allowing it to render images using CPUs and GPUs while simultaneously retaining compatibility with modern optimization libraries~\cite{deepmind2020jax}, deep learning frameworks~\cite{kidger2021equinox}, probabilistic programming languages~\cite{dillon2017tensorflow} and posterior sampling libraries~\cite{blackjax2020github}.
Moreover, we can auto-vectorize our renderer to render multiple images in batches.
This is a useful feature when sampling multiple MCMC chains or when training deep neural networks with stochastic gradient descent (SGD).
Our renderer uses the Phong reflection model~\cite{phong1975illumination} and can render a simple scene with a sphere and a single light source with an image size of $480 \times 640$ in 2.8ms using a low end GPU (GTX 1650 Ti), and in 12ms under the CPU backend (i7-10750H).
Finally, we can use the automatic differentiation system to compute gradients and Hessians for gradient descent optimization or Hamiltonian Monte Carlo (HMC).
Figure~\ref{fig:model_based_optimization2} shows an image rendered with our ray tracer as well as the available differentiable parameters.
We opted to build a flexible system rather than relying on more realistic renderers that use global illumination~\cite{nimier2019mitsuba}.
This flexibility allowed us to directly use within our few-shot learning algorithm, gradient descent optimization followed by posterior sampling with a deep neural network in our likelihood function.

\section{Scene Optimization}\label{sec:scene_optimization}
\begin{figure}[t]
    \begin{subfigure}[T]{0.5\linewidth}
        \begin{subfigure}{\linewidth}
            \caption{\fsclvr}
            \input{sections/images/scene_optimization_fsclvr}~\label{fig:opti_simple}
        \end{subfigure}
        \\ 
        \begin{subfigure}{\linewidth}
            \caption{\fsclvr~optimized train-set materials}
            \includegraphics[width=\linewidth]{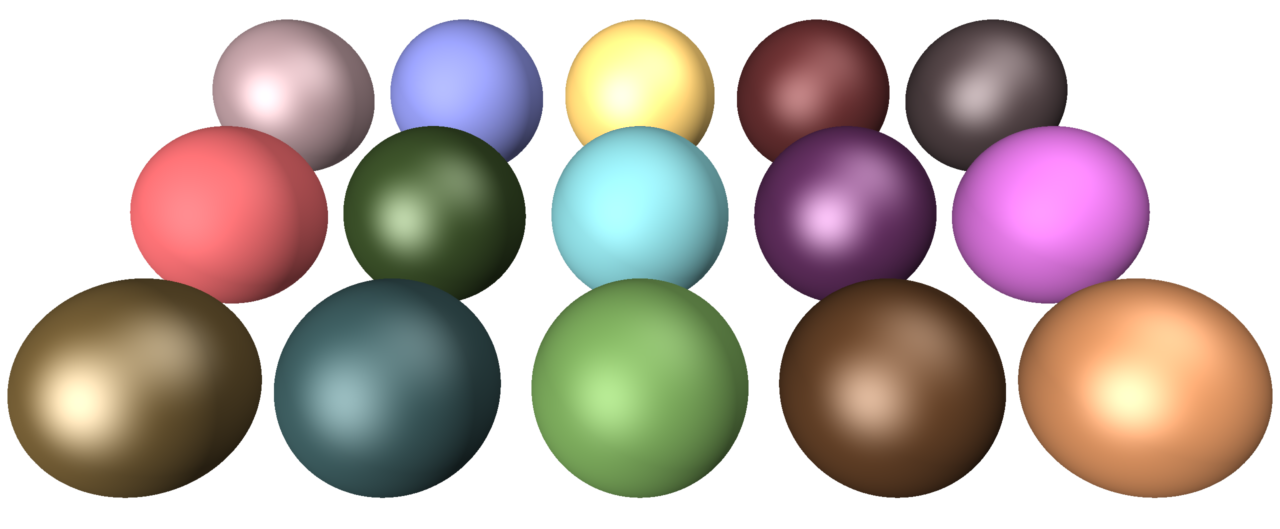}~\label{fig:optimized_materials}
        \end{subfigure}
    \end{subfigure}
    \begin{subfigure}[T]{0.5\linewidth}
        \begin{subfigure}{\linewidth}
            \caption{\fsclvrdark}
            \input{sections/images/scene_optimization_fsclvr_dark}~\label{fig:opti_dark}
        \end{subfigure}
        \\
        \begin{subfigure}{\linewidth}
            \input{sections/images/scene_optimization_fsclvr_room}
            \caption{\fsclvrroom}~\label{fig:opti_room}
        \end{subfigure}
    \end{subfigure}
    \caption{Model based scene optimization: Subfigure~\ref{fig:optimized_materials} displays in each sphere one extracted material obtained from our optimization process. Moreover, one can observe how each material behaves differently under the same lighting conditions.}~\label{fig:model_based_optimization}
    \vspace{-0.5cm}
\end{figure}

    To maximize realism between our generative model and the training dataset, we optimize the scene parametrization $\Scene=\{\Scene_{\Global},\Scene_{\Object} \}$, including global $\Scene_{\Global}$ and object parameters $\Scene_{\Object}$.
The  global parameters
$\Scene_{\Global}=\{
    \Shift_{k},
    \LightIntensity_{k},
    \Color_{\Floor},
    \Pattern_{\Floor},
    \Ambient_{\Floor},
    \Diffuse_{\Floor},
    \}
$
include the $k$th light position, light intensities, floor color, floor pattern, floor ambient, floor diffuse variable.  
The object parameters
$\Scene_{\Object}=\{
\Color_{j},
\Ambient_{j},
    \Diffuse_{j},
    \Specular_{j},
    \Shininess_{j}
    \}
$
include the $j$th training objects' color, ambient, diffuse, specular, and shininess property. Specifically $\mathbf{x}_k,\mathbf{l}_k,\mathbf{c}_F,\mathbf{c}_F, \mathbf{c}_j\in\mathbb{R}^{3}, \mathbf{p}_f\in\mathbb{R}^{H\times W\times 3}$ and all remaining variables are scalars.
To match real and rendered images, for all $N_c$ channels and all $N_p$ pixels we minimize the $\mathrm{L}_{2}$ loss over $\Scene$ using gradient descent through the differentiable renderer $\mathcal{R}(\Scene)$:
\begin{align}
\Scene^{\ast}=\argmin\limits_{\Scene}
\mathrm{L}_2(\mathcal{I}_D,\mathcal{R}(\Scene)).
\end{align}
Moreover, the reflected colors from a realistic material depend on the lighting conditions. 
This implies that color reflections contain information about light location and intensity.
Thus, we optimized the scene parameters by alternating gradient updates between $\Scene_{\Global}$ and $\Scene_{\Object}$ with $N=7$ epochs, $k=5$, and a pattern image of shape $200 \times 200 \times 3$.
We used ADAM with a learning rate of $0.01$.
Figure~\ref{fig:model_based_optimization} shows some results of our optimization problem at different epochs.
In total we optimized 120K pysically interpretable parameters. 
All material point estimates optimized on the \fsclvr~training set are displayed in Figure~\ref{fig:optimized_materials}.

\section{Neural Likelihood}
    While maximizing realism does close the simulation-reality gap, there are still elements for which our differentiable renderer is unable to match perfectly with the true images.
These elements are related to soft shadows, material reflections and different shape topologies.
Thus, we propose a neural metric that measures image discrepancy in feature space.
We apply this neural metric as a likelihood function when fitting a probabilistic graphical model using MCMC.
Using a neural network within MCMC implies performing a forward pass for every sample; moreover, not all featuremaps are relevant for all tasks.
Thus, we opted to use only those feature maps $\Featuremap_m$, where each channel $m$ remains invariant between our rendered images $\Render$ and the true images $\ImageData$
\begin{equation}
    \Featuremap_{m}(\Render) \approx \Featuremap_{m}(\ImageData).
\end{equation}
This implies that we use only those neural features that map similar images to similar feature values.
To construct this invariant transformation we compute the mean square error (MSE) across all feature maps channels $\FeatureChannel$ of VGG16~\cite{simonyan2014very}.
We choose this network due to both its fast inference time, and its widespread application as a feature extraction model~\cite{zhang2018unreasonable}.
Moreover, our methodology is generic and can be applied to any deep learning model.
Specifically, we used the training pairs of true images and the previously optimized image scenes $(\ImageData, \Render^{*}) \in \ImagePairSet$ to compute the accumulated loss $\mathcal{L}(m)$, the channel specific MSE, for each feature map channel $\FeatureChannel$:
\begin{equation}
    \mathcal{L}(m) = \frac{1}{\FeaturemapSize(m)} \sum_{(\ImageData, \Render^{*}) \in \ImagePairSet} \vert \Featuremap_{m}(\Render^{*}) - \Featuremap_{m}(\ImageData) \vert^2.
\end{equation}
where $\FeaturemapSize(m)$ indicates the number of pixel features in channel $m$.
We select the feature channels $\FeatureChannel^{*}$ that accumulate the least amount of loss across the training image pairs $\ImagePairSet$.
For the \fsclvr~training the top 3 channels correspond to the 25th, 51th and 2nd of the 1st convolution layer.
This result is computationally favorable since we can extract the first layer of our model and use it during sampling.

During posterior sampling the neural likelihood takes as input samples from our prior distributions.
The variables associated with those distributions include the object's position $\mathbf{x}\in\mathbb{R}^{3\times 1}$, its angle alongside the z-axis $\theta$, its scales $\mathbf{s}\in\mathbb{R}^{3\times 1}$, its class $\kappa\in\{\text{sphere,cube,cylinder}\}$ and its set of material properties $\Scene_{\Material}=\{\mathbf{c},k^a,k^d,k^s,\alpha\}$.
With these parameters we build the parameterization $\ProbabilisticParameters = \{ \Shift, \Angle, \Scale, 
\Scene_{\Material},\Classes \}$ and formulate the neural likelihood function $\NeuralLikelihood(\ProbabilisticParameters,\Observation_D)$ as:
\begin{equation}~\label{eq:neural_likelihood}
    \NeuralLikelihood(\Observation | \ProbabilisticParameters) \propto
    \prod_{m=1}^{\OptimalFeatureChannelSet^{*}} \prod_{n=1}^{\FeaturemapSize(m)}
    \exp(
        \Featuremap_{m}^{n}(\Observation_D)) -
        \Featuremap_{m}^{n}(\Render(\ProbabilisticParameters)).
\end{equation}

\section{Probabilistic Inverse Graphics Models}\label{sec:probabilistic_inverse_graphics_model}
\subsection{Building Priors}
We provide our inverse graphical model with physically consistent priors for $\Omega$ by considering the following elements:
\begin{itemize}
    \item[$\bullet$] The object translation prior $p(\Shift)$ is a truncated Gaussian distribution ($\mathrm{Tr}\mathcal{N}$) with its limits determined such that the translation of an object has the highest probability mass in the middle of the image plane and 0 mass outside.
    \item[$\bullet$] The angle prior $p(\Angle)$ across the object's the z-axis was selected using a von Mises distribution which assigns equal probability to all angles.
    \item[$\bullet$] The object scales' prior $p(\mathbf{s})$ are modelled as a Log-normal ($\mathrm{Log}\mathcal{N}$) distribution excluding negative scales, or extremely small or large objects.
    \item[$\bullet$] The prior distribution $p(\Classes)$ for the shape classes $\{\mathrm{sphere}, \mathrm{cube}, \mathrm{cylinder} \}$ is a Gumbel-Softmax distribution ($\mathrm{Gmb}$)~\cite{jang2016categorical}, which locates most of the probability mass as one-hot vectors instead of uniformly distributed classes, modelling that objects should not be simultaneously multiple classes at once.
    \item[$\bullet$] We used the optimized parameters from our scene optimization pipeline to build priors for our material variables $\Scene_\Material$.
    Specifically, we fitted a Gaussian Mixture Model (GMM) to each variable using expectation maximization (EM).
    We used 2 mixture components for each variable, and a diagonal covariance matrix.
\end{itemize}
The parameters and distributions of all our priors are shown in the supplementary Section~\ref{sec:supp_priors}.
All GMM prior models are shown in the supplementary Section~\ref{sec:data_driven_priors_results}.

    \subsection{Probabilistic Generative Model}
\begin{figure}[t!]
    \begin{subfigure}[b]{0.60\linewidth}
        \centering
        \begin{tikzpicture}[scale=0.55, transform shape, node distance=0.4cm]

    \node[obs] (image) {$\Observation_{G}$};
    \node[det, above=0.25 of image] (render) {$\Renderer$} ; %
    \node[latent, above=0.25 of render] (object) {$\GenerativeObject$};

    \node[latent, above=0.5 of object, xshift=-3.5cm] (material) {$\GenerativeMaterial$};
    \node[latent, above=0.5 of object, xshift=3.5cm]  (transform) {$\GenerativeTransform$};
    \node[latent, above=of object, yshift=1.0cm]  (shapearg) {$\Classes$};

    \node[latent, above=0.5 of material, xshift=-2cm]  (color) {$\ScalarColor$};
    
    \node[const, above=of color, xshift=-0.5cm]  (empty) {$$}; 
    
    \node[latent, above=0.5 of material, xshift=-1cm]  (ambient) {$\Ambient$};    
    \node[latent, above=0.5 of material, xshift=-0cm]  (specular) {$\Specular$};
    \node[latent, above=0.5 of material, xshift=1cm]  (diffuse) {$\Diffuse$};
    \node[latent, above=0.5 of material, xshift=2cm]  (shininess) {$\Shininess$};

    \node[latent, above=0.5 of transform, xshift=-2.0cm]  (translation) {$t$};
    \node[latent, above=0.5 of transform, xshift=0.0cm]  (rotation) {$\theta$};
    \node[latent, above=0.5 of transform, xshift=2.5cm]  (scale) {$\Scale$};

    \node[const, above=of translation, xshift=-0.5cm] (mean_translation) {$\mu_t$};
    \node[const, above=of translation, xshift=0.5cm]  (variance_translation) {$\sigma_t$}; 
        
    \node[const, above=of rotation, xshift=-0.5cm] (mean_rotation) {$\mu_\theta$};
    \node[const, above=of rotation, xshift=0.5cm]  (variance_rotation) {$\alpha_\theta$}; 

    \node[const, above=of scale, xshift=-0.5cm] (mean_scale) {$\mu_s$};
    \node[const, above=of scale, xshift=0.5cm]  (variance_scale) {$\sigma_s$}; 

    \node[const, above=of shapearg, xshift=-0.5cm]  (count) {$\bm{\sigma}$}; 
    \node[const, above=of shapearg, xshift=0.5cm]  (temperature) {$t$}; 

    \node[const, right=of render]  (camera) {camera}; 
    \node[const, above=of camera]  (lights) {lights}; 
    \node[const, below=of camera]  (background) {ground}; 
 
    \factor[above=of translation, yshift=0.15cm] {translation-f} {left:$\mathrm{Tr}\mathcal{N}$}
        {mean_translation, variance_translation} {translation} ;

    \factor[above=of rotation, yshift=0.25cm] {rotation-f} {left:Mises}
        {mean_rotation, variance_rotation} {rotation} ;

    \factor[above=of scale, yshift=0.10cm] {scale-f} {left:$\mathrm{Log}\mathcal{N}$} {mean_scale, variance_scale} {scale} ;

    \factor[above=of shapearg, yshift=0.25cm] {shapearg-f} {left:$\mathrm{Gmb}$} {count} {shapearg} ;
    \factor[above=of shapearg, yshift=0.25cm] {shapearg-ff} {} {temperature} {shapearg} ;

    \factor[above=of ambient, yshift=0.25cm] {ambient-f} {above:GMM}
        {} {ambient} ;

    \factor[above=of diffuse, yshift=0.25cm] {diffuse-f} {above:GMM}
        {} {diffuse} ;
        
    \factor[above=of specular, yshift=0.25cm] {specular-f} {above:GMM}
        {} {specular} ;

    \factor[above=of shininess, yshift=0.25cm] {shininess-f} {above:GMM}
        {} {shininess} ;

    \factor[above=of color, yshift=0.25cm] {color-f} {above:GMM}
        {} {color} ;

    \edge[-] {shapearg} {object} ;
    \edge[-] {shapearg, material, transform} {object} ;
    \edge[-] {ambient, diffuse, color, specular, shininess} {material} ; 
    \edge[-] {translation, rotation, scale} {transform} ; 
    \edge[-] {object} {render} ; 
    \edge[-] {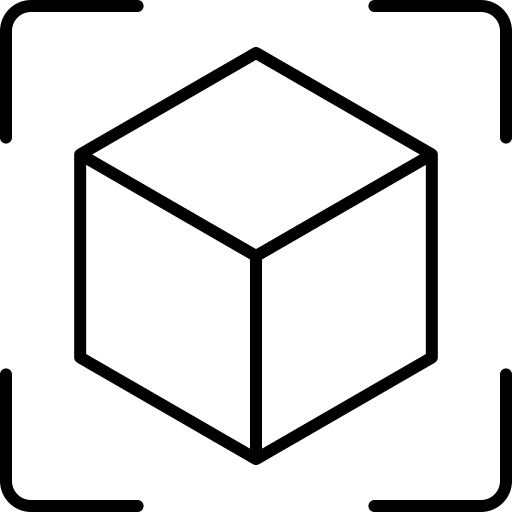} {image} ; 
    \edge[-] {lights} {render} ; 
    \edge[-] {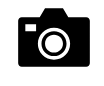} {render} ; 
    \edge[-] {background} {render} ; 

    \plate {colors} {(color)(color-f)(empty)} {$\{r, g, b\}$};
    \plate {translations} {(translation)(translation-f)(mean_translation)(variance_translation)} {$\{x, y\}$};
    \plate {scales} {(scale)(scale-f)} {$\{ s_x, y_y, s_z\}$};

\end{tikzpicture}
        \caption{Probabilistic inverse graphics model}~\label{fig:probabilistic_program}
    \end{subfigure}
    \begin{subfigure}[b]{0.40\linewidth}
        \centering
        \begin{subfigure}[b]{0.235\linewidth}
            \includegraphics[width=\linewidth]{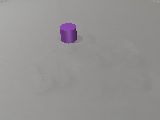}
        \end{subfigure}
        \begin{subfigure}[b]{0.235\linewidth}
            \includegraphics[width=\linewidth]{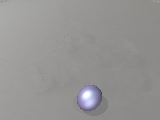}
        \end{subfigure}
        \begin{subfigure}[b]{0.235\linewidth}
            \includegraphics[width=\linewidth]{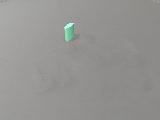}
        \end{subfigure}
        \begin{subfigure}[b]{0.235\linewidth}
            \includegraphics[width=\linewidth]{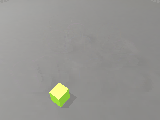}
        \end{subfigure}

        \begin{subfigure}[b]{0.235\linewidth}
            \includegraphics[width=\linewidth]{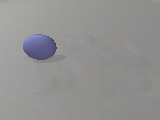}
        \end{subfigure}
        \begin{subfigure}[b]{0.235\linewidth}
            \includegraphics[width=\linewidth]{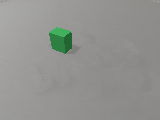}
        \end{subfigure}
        \begin{subfigure}[b]{0.235\linewidth}
            \includegraphics[width=\linewidth]{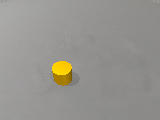}
        \end{subfigure}
        \begin{subfigure}[b]{0.235\linewidth}
            \includegraphics[width=\linewidth]{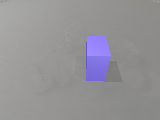}
        \end{subfigure}

        \begin{subfigure}[b]{0.235\linewidth}
            \includegraphics[width=\linewidth]{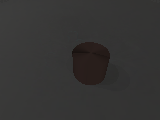}
        \end{subfigure}
        \begin{subfigure}[b]{0.235\linewidth}
            \includegraphics[width=\linewidth]{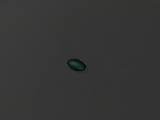}
        \end{subfigure}
        \begin{subfigure}[b]{0.235\linewidth}
            \includegraphics[width=\linewidth]{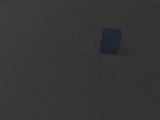}
        \end{subfigure}
        \begin{subfigure}[b]{0.235\linewidth}
            \includegraphics[width=\linewidth]{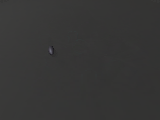}
        \end{subfigure}

        \begin{subfigure}[b]{0.235\textwidth}
            \includegraphics[width=\textwidth]{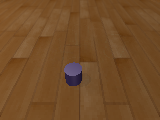}
        \end{subfigure}
        \begin{subfigure}[b]{0.235\linewidth}
            \includegraphics[width=\textwidth]{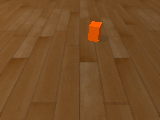}
        \end{subfigure}
        \begin{subfigure}[b]{0.235\linewidth}
            \includegraphics[width=\textwidth]{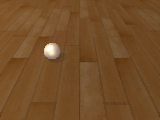}
        \end{subfigure}
        \begin{subfigure}[b]{0.235\linewidth}
            \includegraphics[width=\linewidth]{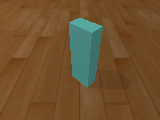}
        \end{subfigure}
        \caption{Prior predictive samples}~\label{fig:prior_predictive_samples}
    \end{subfigure} 
    \caption{Inverse graphics model and prior predictive samples for \fsclvr, \fsclvrdark, and the \fsclvrroom~datasets}~\label{fig:inverse_graphics_model}
    \vspace{-0.70cm}
\end{figure}

Having defined our prior distributions, we proceed to build a probabilistic generative model of images.
Figure~\ref{fig:probabilistic_program} shows our model in plate notation and Figure~\ref{fig:prior_predictive_samples} shows its prior predictive samples.
These prior predictive samples reflect what the model expects to see before making any observation.
Moreover, our model outputs a probability for each sample.
This all being referred as the prior predictive distribution (PPD).
We now proceed to explain in detail the probability density functions (PDFs) and the deterministic functions applied to our model.

We define the (PDF) over possible materials $\Material$ by considering each material property in $\Scene_{\Object}$ 
as independent:
\begin{equation}\label{eq:pdf_material}
    p(\Material|\Scene_{\Material}
    ) =
        \prod_{m \in \Scene_{\Material}} \left[ \sum_{k=0}^{K} \pi_{k}^{m} \mathcal{N}(m | \mu_k^{m}, \Sigma_{k}^{m}) \right].
\end{equation}
As shown in Figure~\ref{fig:probabilistic_program} the PDF over possible affine transformations $\GenerativeTransform$ includes a $\TruncatedNormalDistribution$, von Mises, and a $\LogNormalDistribution$~distribution, and considers the translation, angle and scale as independent events:
\begin{equation}~\label{eq:pdf_transform}
   p(\GenerativeTransform | \Shift, \Angle, \Scale) = 
   \prod_{t \in T} \left[ \frac{\TruncatedNormalDistribution(\Shift | \mu_t, \sigma_t)}{2\pi I_0(0)} \right]
   \prod_{s \in S} \frac{ \exp(-\frac{(\ln(s) - \mu_{s})^2}{2 \sigma_{s}^{2}} ) }{s \sigma_{s} \sqrt{2 \pi}}.
\end{equation}
where $T = \{x, y \}$, $S = \{ s_x, s_y, s_z \}$ and $I_0$ is the Bessel function of order zero.
Finally, we compute the probability density function over objects $p(\GenerativeObject | \Material, \GenerativeShape, \GenerativeTransform) $ using the previously defined PDFs for materials~\ref{eq:pdf_material} and transforms~\ref{eq:pdf_transform}, as well as $p(\Classes  | \bm{\sigma}, \mathrm{t})$ corresponding to the Gumbel-Softmax probability with a class probability $\bm{\sigma}$ and a temperature $t$
\begin{equation}
    p(\Classes | \bm{\sigma}, \mathrm{t})
    p(\Material | \Scene_{\Material}
    )
    p(\GenerativeTransform | \Shift, \Angle, \Scale).
\end{equation}
Furthermore, the object variable $\GenerativeObject$ is passed through our deterministic differentiable rendering function $\DiffGraphics$ outputting image samples, and similarly to~\cite{gothoskar20213dp3} we build an affine transform that enforces objects to not collide with the floor.
Having defined a forward generative model of images, we now perform Bayesian inference in order to obtain the posterior distributions over the variables $\ProbabilisticParameters$ given an observation $\Observation$, here denoting a test image.

\begin{equation}
    p(\ProbabilisticParameters | \Observation) \sim p(\Observation | \ProbabilisticParameters) p(\ProbabilisticParameters)\label{eqn:target_probability}.
\end{equation}
We define a color likelihood function $\ColorLikelihood$ using a truncated normal distribution over each pixel argument $(u, v, c)$ of an image $\Observation$ with shape [H, W, 3]:
\begin{equation}
\xi(u, v, c) = \Render(u, v, c | \ProbabilisticParameters) - \ImageData(u, v, c)
\end{equation}
\begin{equation}~\label{eq:color_likelihood}
    \Likelihood(u, v | \ProbabilisticParameters)_{\mathrm{C}} =
        \prod_{u=0}^{W} \prod_{v=0}^{H} \prod_{c=0}^{3} \mathrm{Tr}\mathcal{N}(\xi(u, v, c), \sigma_{I}). \\
\end{equation}
All parameters of our PGM are displayed in the supplementary Section~\ref{sec:supp_priors}.
Given our target probability function, and an observation $\Observation$, we can now sample from a conditioned posterior distribution using MCMC. \\

    \subsection{Posterior Sampling}
    \begin{figure}[t!]
    \centering
    \begin{subfigure}[T]{0.48\linewidth}
        \centering
        \includegraphics[width=\textwidth]{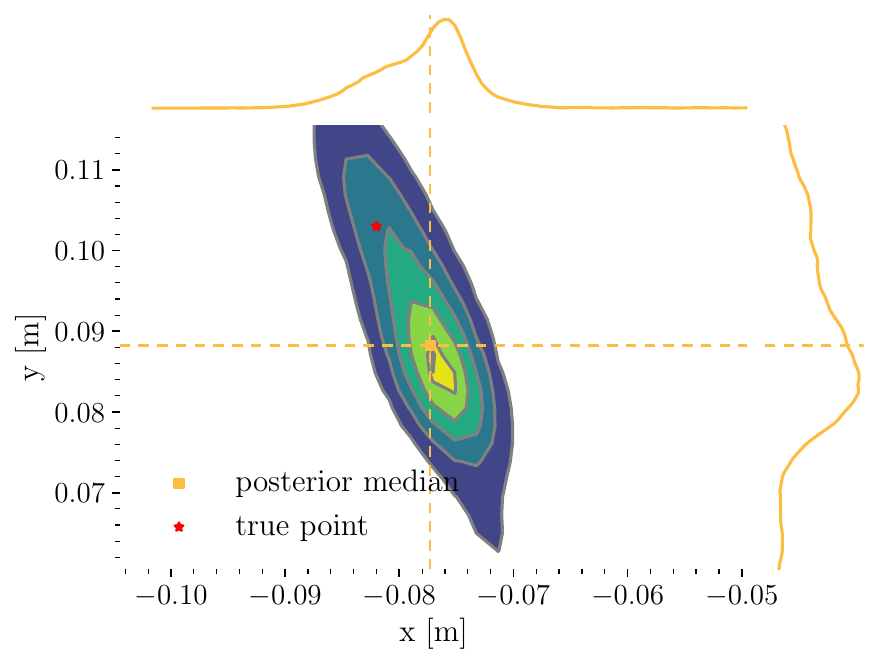}
        \caption{Translation $\Shift$ posterior}~\label{fig:shift_posterior}
    \end{subfigure}
    \begin{subfigure}[T]{0.48\linewidth}
        \includegraphics[width=0.95\textwidth]{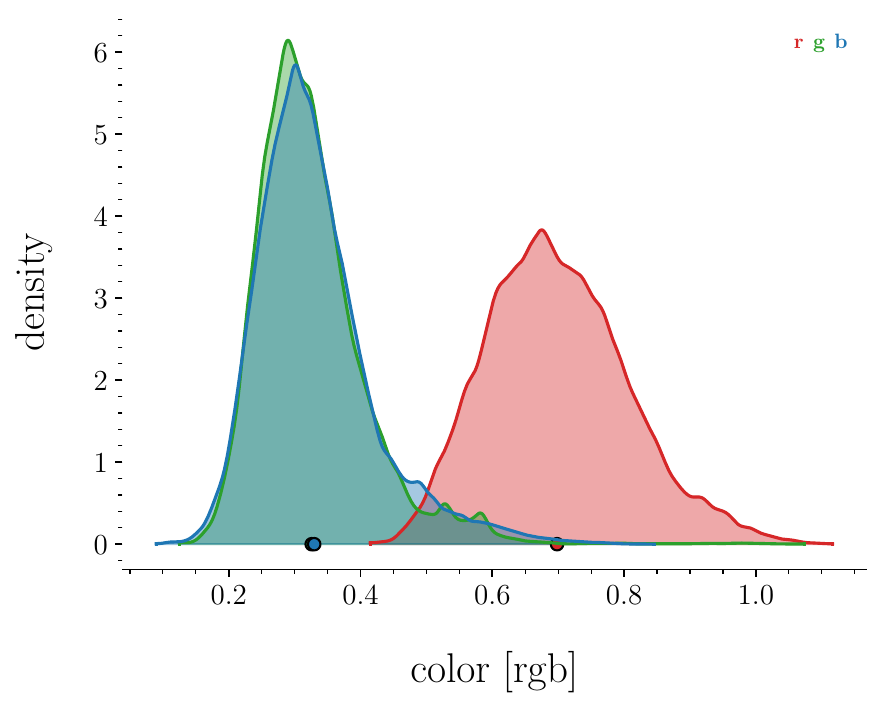}
        \caption{Color $\Color$ posterior}~\label{fig:color_posterior}
    \end{subfigure}
    \caption{Posteriors from a single test image. Each contour in~\subref{fig:shift_posterior}) indicates the highest density intervals with 5\%, 10\%, 20\%, 40\%, 60\%, 80\% probability}~\label{fig:posterior_results}
    \vspace{-0.80cm}
\end{figure}
For each of the test images we sampled from the target distribution using the Rosenbluth-Metropolis Hastings (RMH) MCMC method. 
We used 20 chains sampling 30K posterior samples per chain with a burn-in of 1K samples.
We used a diagonal Gaussian kernel as our proposal generator which we initialized with all diagonal elements being $0.05$.
Before sampling we performed automatic tuning of our chains to have an acceptance rate between $20\%$ and $50\%$.
In order to speed-up sampling we reparametrized our prior distributions and resized each test image to a shape of $ \left[ 160, 120, 3 \right] $, and removed all shadows from our ray-tracing pipeline.
Details of the prior reparametrization are show in the supplementary Section~\ref{sec:bijection_parameters}.
In total we approximated 420 13-dimensional integrals using 252M samples in approximately 14 hours using an low budget RTX A4000 GPU.
Figure~\ref{fig:posterior_results} shows the posterior distributions over 2 variables after conditioning on a single test image.
The posterior distribution of the translation components $\Shift$ is shown in Figure~\ref{fig:shift_posterior}.
This posterior indicates more uncertainty across the y-axis.
This result is seen across most of our samples, and seems logical given that the y-axis provides us less pixel information about the possible position and size of the objects.
Moreover, while the point estimate of our translation is off by one centimeter in the y-axis, the posterior distribution captures the direction of the biggest source of error.
This result can be relevant for robot manipulation, where we could inform the robot's behavior with the given uncertainty in order to consider it while performing a task.
Additional posterior results and point estimates can be observed in the supplementary Section~\ref{sec:posterior_results}.

\begin{figure}[t!]
    \centering
    \begin{subfigure}{0.32\linewidth}
    \includegraphics[width=\textwidth]{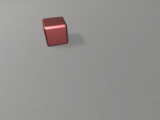}
    \caption{Observation}~\label{fig:true_test_image}
    \end{subfigure}
    \begin{subfigure}{0.32\linewidth}
        \includegraphics[width=\textwidth]{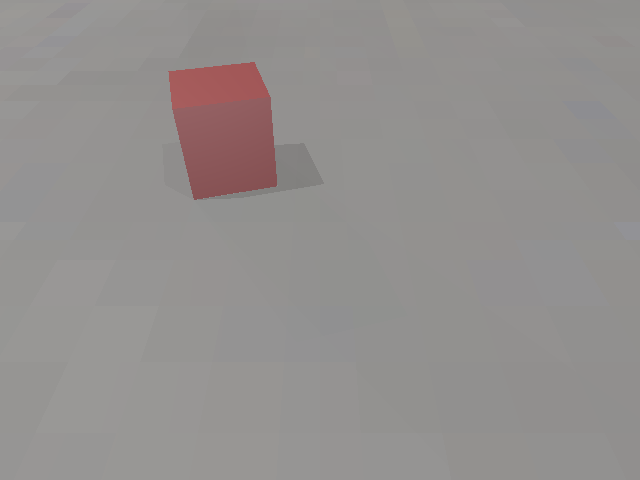}
        \caption{Median point estimate}~\label{fig:short-b}
    \end{subfigure}
    \begin{subfigure}{0.32\linewidth}
        \includegraphics[width=\textwidth]{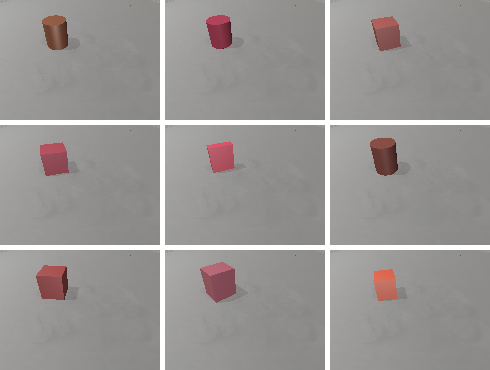}
        \caption{Posterior samples}~\label{fig:posterior_samples}
    \end{subfigure}
    \caption{Posterior results from conditioning on a single observation: Subfigure~\ref{fig:true_test_image} shows the observation used for conditioning our PGM.
    Subfigure~\ref{fig:short-b} shows a single point summary of the estimated posterior distributions of the PGM after conditioning with the observation.
    We can observe how this point estimate provides an accurate description of the observation.
    Moreover, we are not only estimating a single point value for each of our variables, but rather full probability distributions.
    Subfigure~\ref{fig:posterior_samples} shows how our posterior distributions capture similar representations of the concept seen in the observation.}~\label{fig:posterior_results2}
    \vspace{-0.50cm}
\end{figure}

\section{Probabilistic Prototypical Programs}\label{sec:probabilistic_prototypical_programs}
    In this section we reuse the posterior samples to build a prototypical object representation.
Specifically, we perform a Gaussian kernel density estimate using Scott's bandwidth~\cite{scott1992density} for the posterior samples of the variables in $\Scene_{\mathcal{S}}=\{\Scale,
\Scene_{\Material},\Classes\}$.
Thus, our prototypical representation consists of all our previous posterior samples excluding the pose variables $\Shift$ and $\theta$.
These posterior distributions are used to build a new generative model by removing these variables from our previous inverse graphics model~\ref{fig:probabilistic_program}.
A visualization of this new graphical model is shown in the supplementary Section~\ref{sec:supplementary_prototypical}.
Moreover, we define the functions \distance~and \merge~between two generative models.
We define the \textit{distance} between two generative models $C_I$ and $C_k$ as the sum of the Kullback-Leibler divergences across our selected variables $\Scene_{\mathcal{S}}$ 
\begin{equation}
    d(C_{I}, C_{k}) =
        \sum_{\ProtoVariables} \text{KL}(C_{k}^{\ProtoVariables} \| C_{I}^{\ProtoVariables}) =
        \sum_{\ProtoVariables} \sum_{x} C_{k}^{\ProtoVariables}(x) \log \frac{C_{k}^{\ProtoVariables}(x)}{C_{I}^{\ProtoVariables}(x)}.
\end{equation}
The \merge~operation ($\otimes$) is defined as performing KDE on the concatenated posterior samples for each variable in both PGMs.
We define this conditioned probabilistic generative models with these two additional functions as a protoprogram $C$.
Moreover, protoprograms have the additional functionalities associated to all PGMs; those being the capacity to \texttt{sample}~and to compute the \texttt{probability} of any given sample.
Consequently, protoprograms allow us to generate new concepts and images that are similar to our given prototypical observation.
Figure~\ref{fig:posterior_samples} shows ordered samples of the protoprogram conditioned on observation~\ref{fig:true_test_image}.
This characteristic was indicated to be a relevant human trait that is not present in most few-shot learning methods~\cite{lake2019omniglot}.
Having defined our protoprograms we proceed to present a classification algorithm.
Given $\mathrm{N}$ example images (shots) of an object class $k$, we proceed to build a protoprogram for this class by merging all conditioned programs as $C_k = C_1^{k} \otimes C_{2}^{k} \otimes \ldots C_{\mathrm{N}}^{k}$.
Given a new observation $I$, we proceed to learn a program $C_{I}$ following the same structure as before.
We then use the program \textit{distance} to compute a probability conditioned on each protoprogram class $c$ by applying softmax on the negative program distances.

\begin{equation}
	p(c=k | I) = \frac{\exp(-d(C_{I}, C_{k}))}{\sum_{k} \exp(-d(C_{I}, C_{k}))}.\label{eqn:softmax}
\end{equation}

\section{Results}\label{sec:results}
    We present the results of our prototypical probabilistic programs (P3) applied to the datasets \fsclvr, \fsclvrroom, \fsclvrdark, and \ycbood.
As previously mentioned, these datasets test generalization in a low sample training regime, as well as in different background, lighting and out-of-distribution shapes.
Moreover, we tested two versions of our model.
\begin{table}[h!]
\scriptsize
\caption{Few-shot accuracy results for multiple shots}~\label{tab:few-shot}
 \begin{subtable}{\linewidth}
        \centering
 		\begin{tabular}{l|cccc|cccc}
\hline
\multirow{2}{*}{\textbf{Model}} & \multicolumn{4}{c|}{\underline{\textbf{\fsclvr~test dataset}}} & \multicolumn{4}{c}{\underline{\textbf{\fsclvrroom~test dataset}}} \\
& \multicolumn{2}{c}{\textbf{5-way}} & \multicolumn{2}{c|}{\textbf{10-way}} & \multicolumn{2}{c}{\textbf{5-way}} & \multicolumn{2}{c}{\textbf{10-way}} \\
& \textbf{1-shot} & \textbf{5-shot} & \textbf{1-shot} & \textbf{5-shot} & \textbf{1-shot} & \textbf{5-shot} & \textbf{1-shot} & \textbf{5-shot} \\
\hline
$\textrm{MAML}_{56 \times 56}$~\cite{finn2017model} & 76.96\% & 87.48\% & 62.82\% & 69.84\% & 77.94\% & 79.18\% & 47.23\% & 61.46\% \\
$\textrm{MAML}_{84 \times 84}$~\cite{finn2017model} & 70.86\% & 32.38\% & 51.77\% & 64.95\% & 72.96\% & 81.72\% & 32.12\% & 57.01\% \\
$\textrm{ProtoNets}_{56 \times 56}$~\cite{snell2017prototypical} & 60.52\% & 85.54\% & 51.22\% & 79.73\% & 66.70\% & 83.04\% & 54.65\% & 72.18\% \\
$\textrm{ProtoNets}_{84 \times 84}$~\cite{snell2017prototypical} & 51.16\% & 74.94\% & 40.13\% & 67.09\% & 47.12\% & 68.32\% & 33.65\% & 54.41\% \\
\hline
P3 (ours) & 95.98\% & \textbf{98.98\%} & 92.61\% & \textbf{97.51\%} & 91.66\% & 96.98\% & 86.01\% & 93.59\% \\
NP3 (ours) & \textbf{97.18\%} & 98.86\% & \textbf{94.52\%} & 97.47\% & \textbf{94.36\%} & \textbf{98.96\%} & \textbf{90.52\%} & \textbf{97.53\%} \\
\hline
\end{tabular}
        \caption{Accuracies on the \fsclvr~and \fsclvrroom~test dataset}~\label{tab:results1}
    \end{subtable}
    \hfill 
  \begin{subtable}{\linewidth}
        \centering
        \begin{tabular}{l|cccc|cccc}
\hline
\multirow{2}{*}{\textbf{Model}} & \multicolumn{4}{c|}{\underline{\textbf{\fsclvrdark~test dataset}}} & \multicolumn{4}{c}{\underline{\textbf{\ycbood~test dataset}}} \\
& \multicolumn{2}{c}{\textbf{5-way}} & \multicolumn{2}{c|}{\textbf{10-way}} & \multicolumn{2}{c}{\textbf{5-way}} & \multicolumn{2}{c}{\textbf{10-way}} \\
& \textbf{1-shot} & \textbf{5-shot} & \textbf{1-shot} & \textbf{5-shot} & \textbf{1-shot} & \textbf{5-shot} & \textbf{1-shot} & \textbf{5-shot} \\
\hline
$\textrm{MAML}_{56 \times 56}$~\cite{finn2017model} & 79.56\% & 86.22\% & 60.39\% & 62.77\% & 47.22\% & 32.70\% & 36.55\% & 39.16\% \\
$\textrm{MAML}_{84 \times 84}$~\cite{finn2017model} & 69.08\% & 80.36\% & 49.15\% & 63.00\% & 52.22\% & 37.56\% & 39.83\% & 47.56\% \\
$\textrm{ProtoNets}_{56 \times 56}$~\cite{snell2017prototypical} & 65.34\% & 86.66\% & 52.88\% & 79.32\% & 49.80\% & 67.26\% & 39.13\% & 57.37\% \\
$\textrm{ProtoNets}_{84 \times 84}$~\cite{snell2017prototypical} & 43.92\% & 70.58\% & 32.51\% & 58.19\% & 41.26\% & 62.28\% & 30.96\% & 50.33\% \\
\hline
P3 (ours) & 96.86\% & 98.04\% & 93.88\% & 96.62\% & 66.42\% & 67.84\% & 60.32\% & 57.31\% \\
NP3 (ours) & \textbf{97.30\%} & \textbf{98.48\%} & \textbf{94.88\%} & \textbf{97.31\%} & \textbf{67.44\%} & \textbf{71.26\%} & \textbf{62.09\%} & \textbf{61.36\%} \\
\hline
\end{tabular}

        \caption{Accuracies on the \fsclvrdark~and \ycbood~test dataset}~\label{tab:results2}
    \end{subtable}
\end{table}
\vspace{-1.5cm}
\begin{figure}[h!]
    \centering
    \begin{subfigure}[b]{1.0\textwidth}
        \centering
        \includegraphics[width=0.24\textwidth]{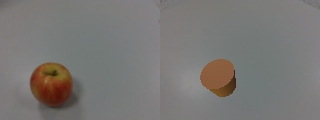}
        \includegraphics[width=0.24\textwidth]{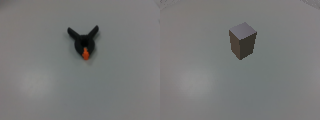}
        \includegraphics[width=0.24\textwidth]{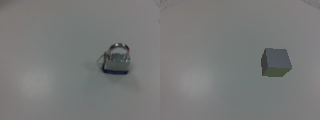}
        \includegraphics[width=0.24\textwidth]{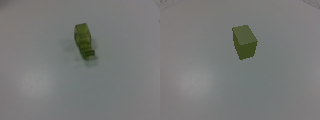}
    \end{subfigure}
    \\
    \begin{subfigure}[b]{1.0\textwidth}
        \centering
        \includegraphics[width=0.24\textwidth]{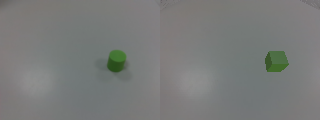}
        \includegraphics[width=0.24\textwidth]{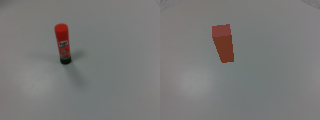}
        \includegraphics[width=0.24\textwidth]{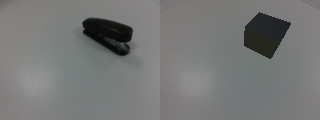}
        \includegraphics[width=0.24\textwidth]{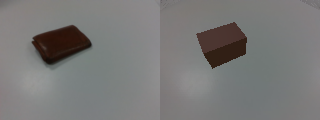}
    \end{subfigure}
    \caption{NP3 applied to real images}~\label{fig:real_images}
    \vspace{-0.5cm}
\end{figure}
These correspond to a generative model which uses both the color and neural likelihood functions (NP3), and the P3 model which only considers the color likelihood.
We compared the performance of our model with respect to the standard metric-based, and meta learning neural-only models in Table~\ref{tab:few-shot}.
These are respectively ProtoNets~\cite{snell2017prototypical}, and MAML~\cite{finn2017model}.
We used the same learning rate (1e-3) and the same resolution ($84 \times 84$)~suggested by the original authors.
Moreover, we observed that sometimes the resolution of $56 \times 56$ was outperforming the $84 \times 84$ resolution; thus we include both results.
In the supplementary Section~\ref{sec:sup_maml} we show more experiments with different resolutions.
Our model outperforms these few-shot neural models while using considerably less parameters.
Our probabilistic graphical model P3 has only 13 parameters per class ($\ProbabilisticParameters$).
That is a 1,738-fold decrease in parameter count when using 5-ways.
While the number of parameters should not be taken as a measure of model complexity, this parameter reduction allows us to perform full Bayesian inference and compute the posterior of our parameters to determine the uncertainty of our model.
Furthermore, our zero-shot pose estimation evaluation can be found in the supplementary Section~\ref{sec:sup_pose}.
Finally, as observed in Figure~\ref{fig:real_images}, without any modification to our model we are able to directly apply it  to unseen real objects.

\section{Conclusion and Future Work}\label{sec:conclusions}
    This paper presents a novel Bayesian inverse graphics framework that encodes images as prototypical probabilistic programs.
This approach addresses some of the limitations of existing neural algorithms such as high sample and model complexity, and lack of uncertainty quantification.
Moreover, we use these probabilistic programs to build a few-shot learning classification algorithm. 
This algorithm integrates our newly introduced differentiable renderer with probabilistic programming languages, gradient descent optimization libraries, and deep learning frameworks.
Our method achieves higher classification accuracy than standard few-shot neural methods, while using considerably less parameters.
Furthermore, we demonstrate generalization to different lighting conditions, backgrounds, and unseen complex objects.
Additionally, we proposed a neural likelihood function which combines deep learning models with Bayesian inference.
In future work we plan to extend our generative models to merge primitive shapes to fit more complex objects.

\vspace{0.75cm}
\\
\textbf{Acknowledgements} \\
    This work was funded by the German Aerospace Center (DLR) with federal funds (Grant 50RA2126A / 50RA2126B) from the German Federal Ministry of Economic Affairs and Climate Action (BMWK) in the project PhysWM.

\bibliographystyle{splncs04}
\bibliography{references}


\section{Physics based priors and likelihood parameters}~\label{sec:supp_priors}
\begin{figure}[htpb]
    \begin{subfigure}[T]{0.29\linewidth}
        \centering
        \includegraphics[width=\linewidth]{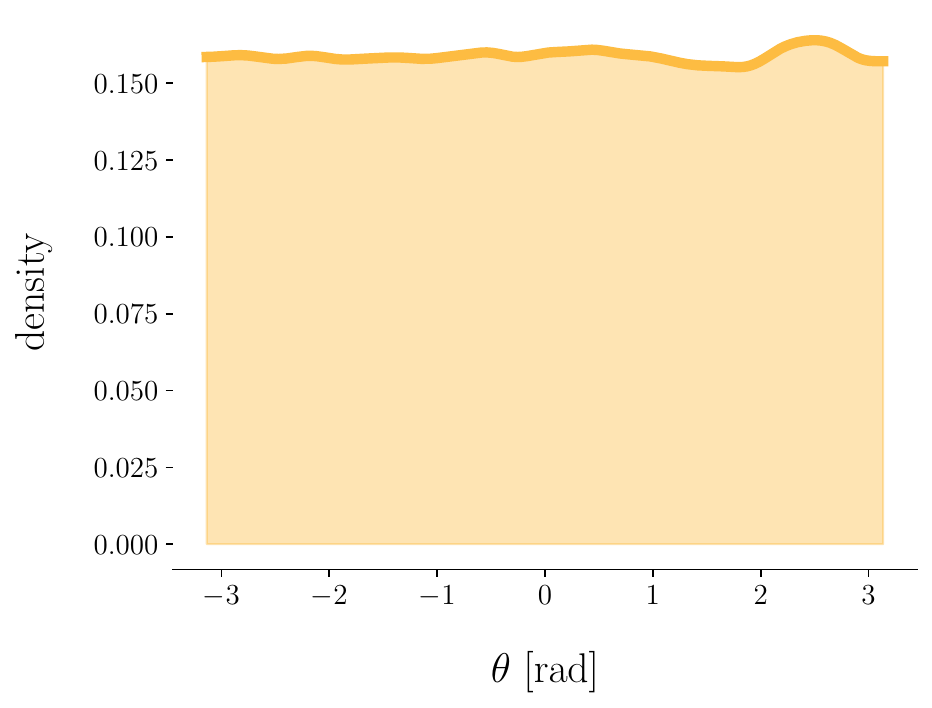}
        \caption{Theta prior}~\label{fig:prior_forward_theta}
    \end{subfigure}
    \begin{subfigure}[T]{0.29\linewidth}
        \centering
        \includegraphics[width=\linewidth]{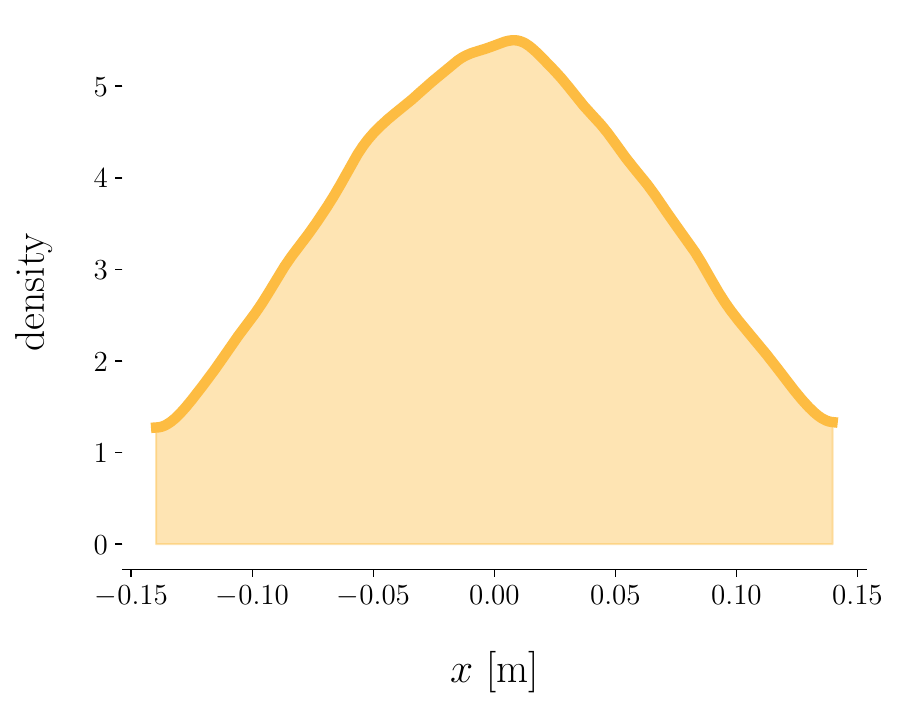}
        \caption{Translation in x prior}~\label{fig:prior_forward_translation_x}
    \end{subfigure}
    \begin{subfigure}[T]{0.29\linewidth}
        \centering
        \includegraphics[width=\linewidth]{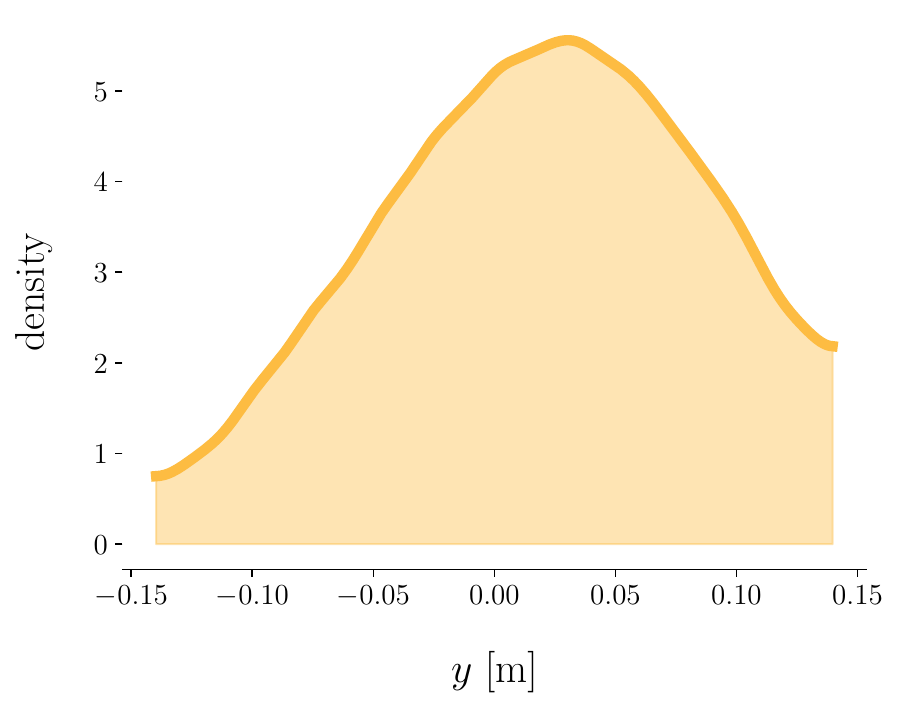}
        \caption{Translation in y prior}~\label{fig:prior_forward_translation_y}
    \end{subfigure}
    \\
    \begin{subfigure}[T]{0.29\linewidth}
        \centering
        \includegraphics[width=\linewidth]{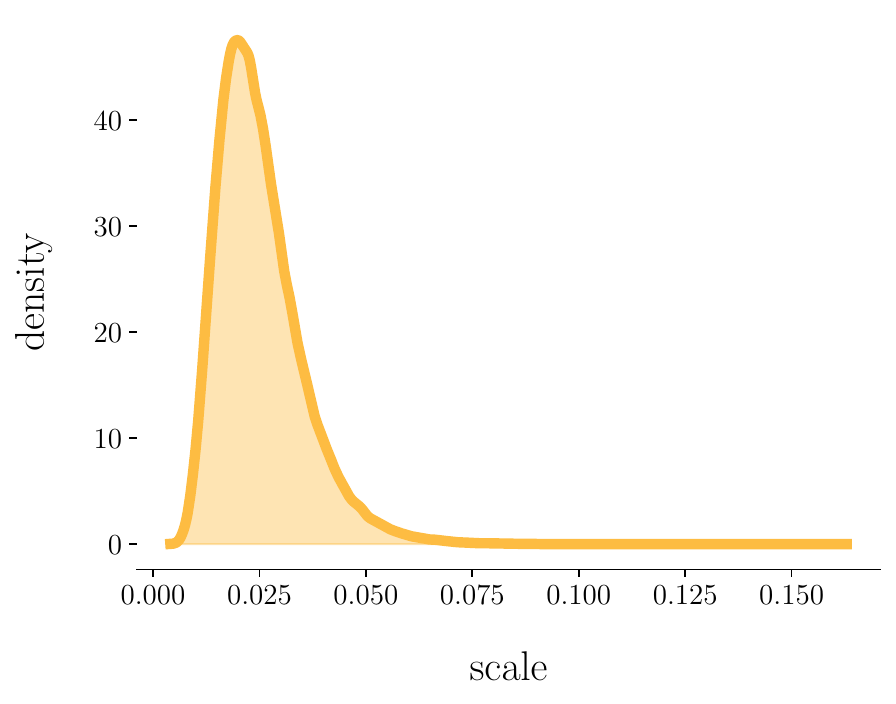}
        \caption{Scale prior}~\label{fig:prior_forward_scale}
    \end{subfigure}
    \begin{subfigure}[T]{0.29\linewidth}
        \centering
        \includegraphics[width=\linewidth]{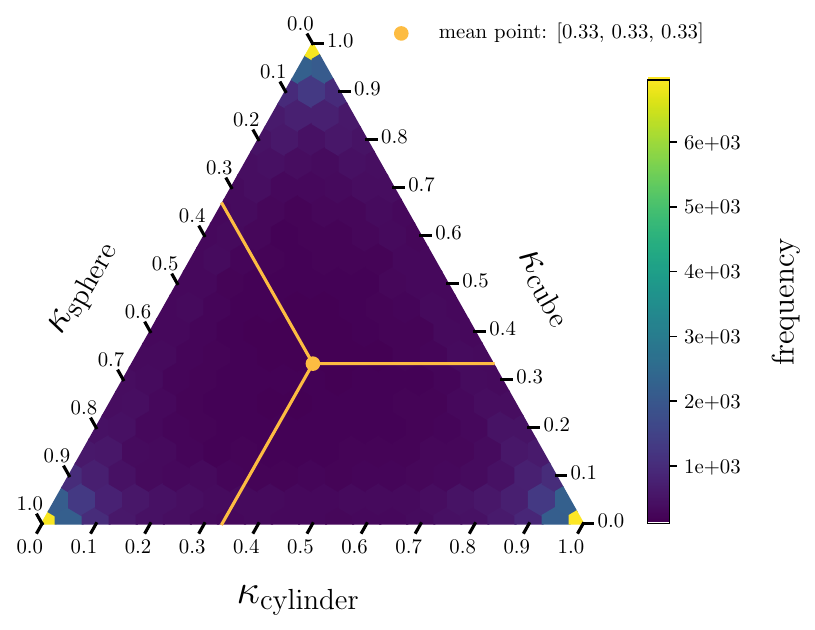}
        \caption{Classes prior}~\label{fig:prior_forward_classes}
    \end{subfigure}
    \caption{Prior distributions}~\label{fig:priors_appendix}
\end{figure}

    The parameters and visualizations of our prior distributions and likelihood functions are displayed in Figure~\ref{fig:priors_appendix} and Table~\ref{tab:prior_parameters} shows all our prior and likelihood parameters.

\vspace{-0.5cm}
\begin{table}[ht]
    \centering
    \caption{Prior and likelihood parameters}~\label{tab:prior_parameters}
    \resizebox{0.50\linewidth}{!}{
    \begin{tabular}{|l|c|}
        \hline
        \textbf{Parameter} & \textbf{Value} \\
        \hline
        \texttt{shift mean} ($\mu_t$) & [0.0, 0.025] \\
        \texttt{shift scale} ($\sigma_t$) & 0.08 \\
        \texttt{theta mean} ($\mu_{\theta}$) & 0.0 \\
        \texttt{theta concentration} ($\alpha_{\theta}$) & 0.0 \\
        \texttt{scale mean} ($\mu_s$) & 0.025 \\
        \texttt{scale scale} ($\sigma_s$) & 0.0001 \\
        \texttt{classes temperature} (t) & 0.5 \\
        \texttt{class probabilities} ($\sigma$) & [1/3, 1/3, 1/3] \\
        \texttt{Neuro likelihood $\propto$ constant} & 0.05 \\
        \texttt{Color likelihood scale ($\sigma_I$)} & 1.0 \\
        \texttt{Color likelihood scale ($\sigma_I^{\text{\tiny{FS-CLVR-DARK}}}$)} & 0.35 \\
        \hline
    \end{tabular}
    }
\end{table}

\section{Training data based priors}~\label{sec:data_driven_priors_results}
    In this section we show the data driven priors fitted for the \fsclvr~dataset for the variables  color $\Color$, ambient $\Ambient$, diffuse $\Diffuse$, specular $\Specular$, and shininess $\Shininess$.

\begin{figure}[htpb]
    \begin{subfigure}[T]{0.23\linewidth}
        \centering
        \includegraphics[width=\linewidth]{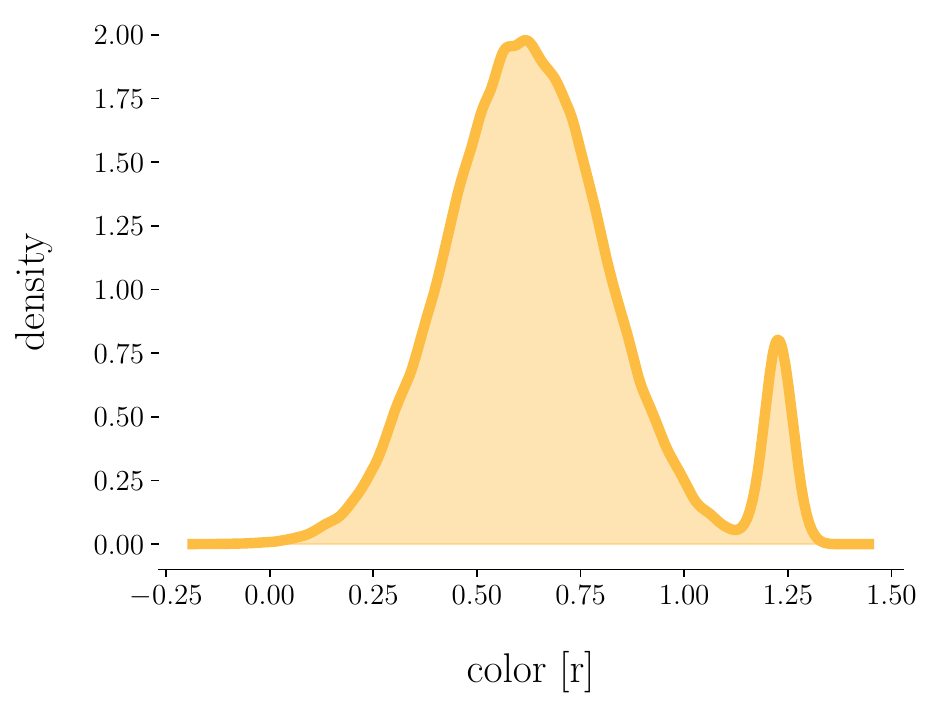}
        \caption{GMM prior for color channel R}~\label{fig:prior_forward_R}
    \end{subfigure}
    \begin{subfigure}[T]{0.23\linewidth}
        \centering
        \includegraphics[width=\linewidth]{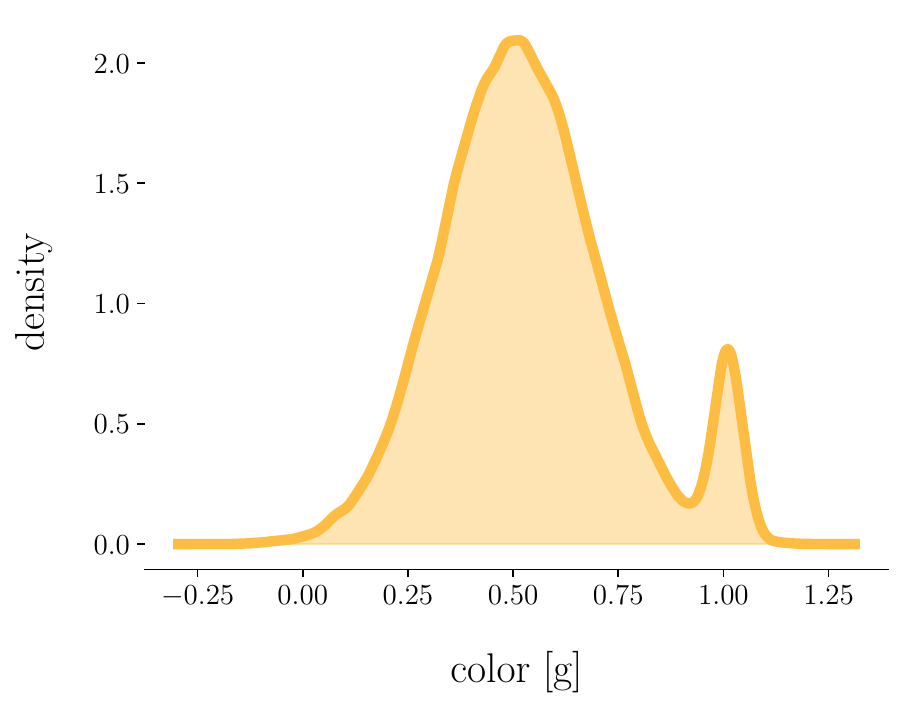}
        \caption{GMM prior for color channel G}~\label{fig:prior_forward_G}
    \end{subfigure}
    \begin{subfigure}[T]{0.23\linewidth}
        \centering
        \includegraphics[width=\linewidth]{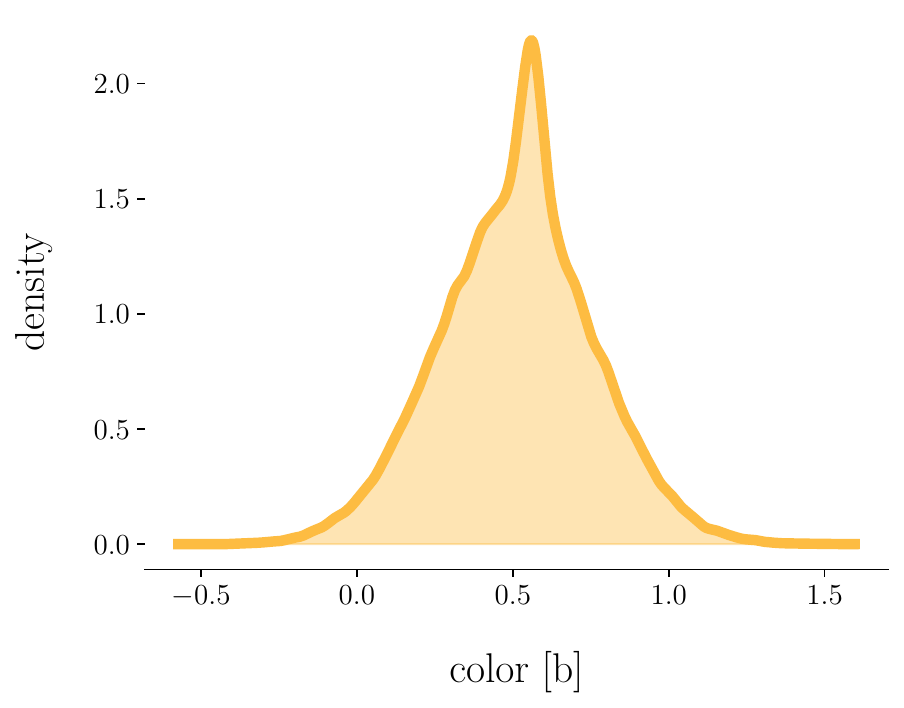}
        \caption{GMM prior for color channel B}~\label{fig:prior_forward_B}
    \end{subfigure}
    \begin{subfigure}[T]{0.23\linewidth}
        \centering
        \includegraphics[width=\linewidth]{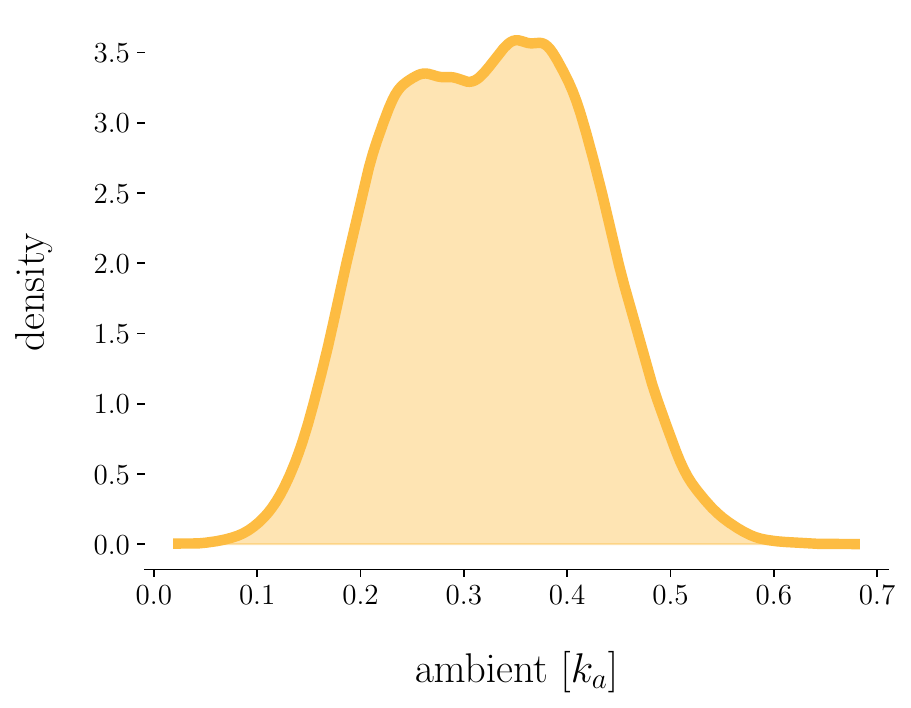}
        \caption{GMM prior for ambient variable}~\label{fig:prior_forward_ambient}
    \end{subfigure}
    \begin{subfigure}[T]{0.23\linewidth}
        \centering
        \includegraphics[width=\linewidth]{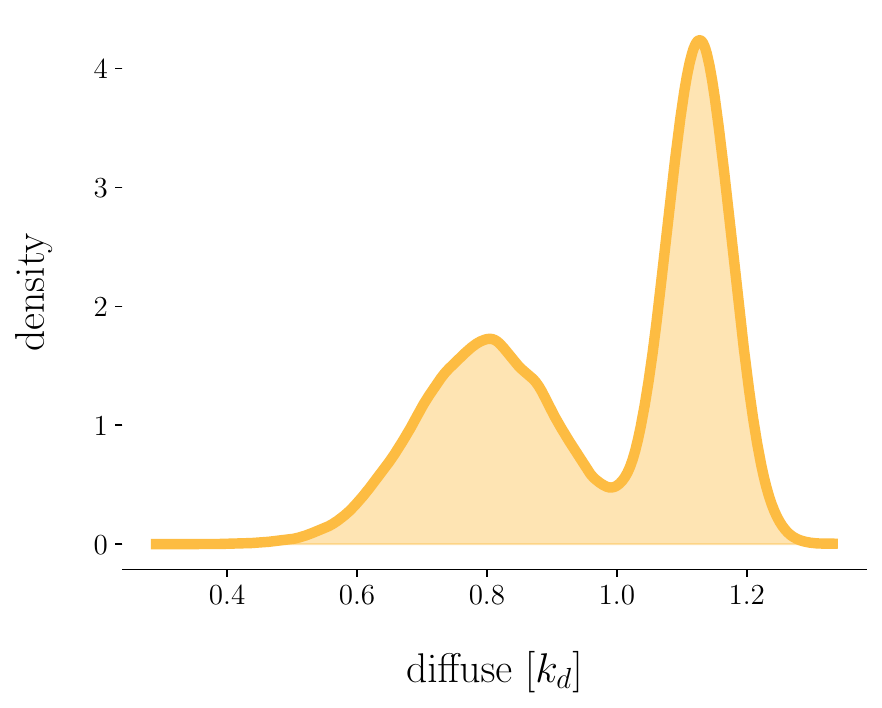}
        \caption{GMM prior for diffuse variable}~\label{fig:prior_forward_diffuse}
    \end{subfigure}
    \begin{subfigure}[T]{0.23\linewidth}
        \centering
        \includegraphics[width=\linewidth]{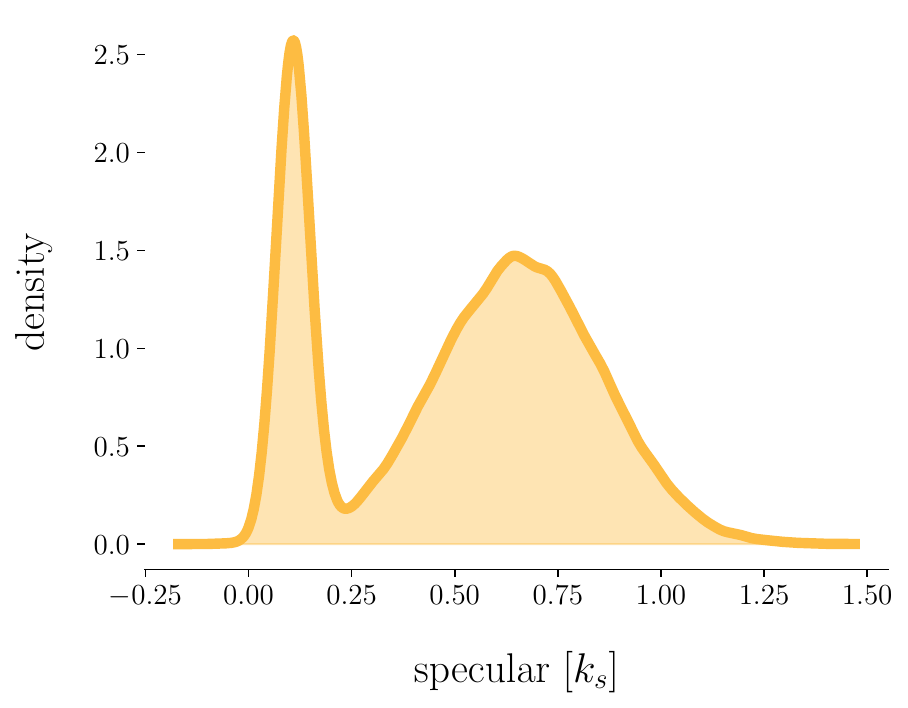}
        \caption{GMM prior for specular variable}~\label{fig:prior_forward_specular}
    \end{subfigure}
    \begin{subfigure}[T]{0.23\linewidth}
        \centering
        \includegraphics[width=\linewidth]{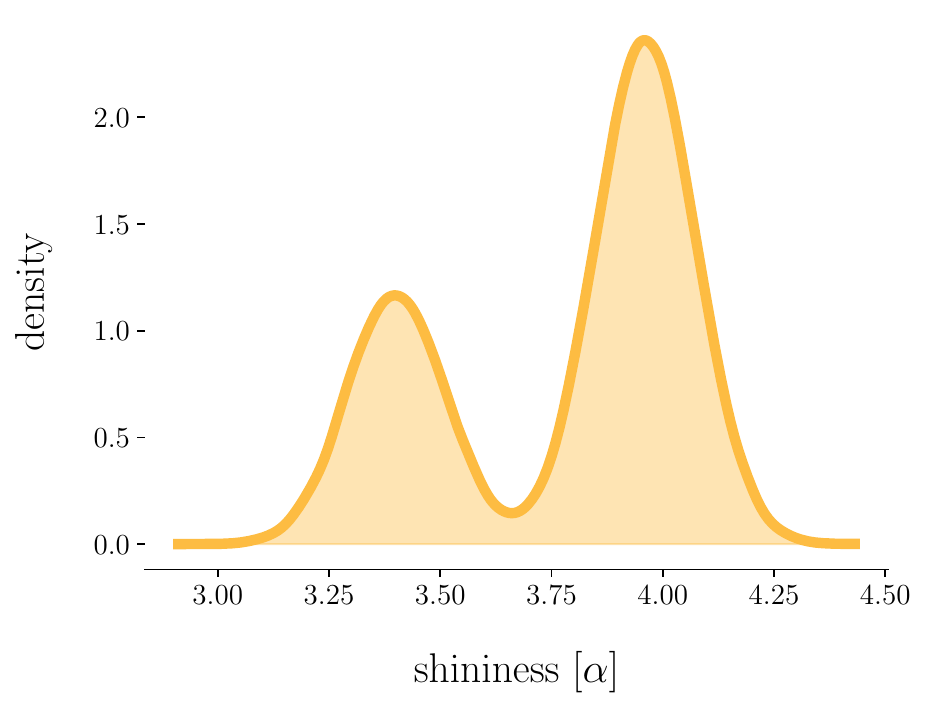}
        \caption{GMM prior for shininess variable}~\label{fig:prior_forward_shininess}
    \end{subfigure}
    \caption{Data-driven priors.}
\end{figure}

\section{Bijection results}~\label{sec:bijection_parameters}
Reparametrizing probability distributions can increase the number of effective samples without increasing computional costs.
We transformed the prior probability distributions to fit Normal distributions using the appropiate bijection functions. 
Specifically for $\Shift$, $\Angle$, $\Scale$, $\Classes$ we apply the following bijections parametrized by $\omega$ and $\phi$:
\begin{align}
    \hat{\Shift}   &= \omega_{\Shift} \Shift + \phi_{\Shift} \\
    \hat{\Angle}   &= \mathrm{sigmoid}(\frac{\pi}{2} \Angle) \\
    \hat{\Scale}   &= \mathrm{sigmoid}(\omega_{\Scale} \Scale + \phi, -\pi, \pi) \\
    \hat{\Classes} &= \mathrm{softmax}(\omega_{\Classes} \Classes)
\end{align}

Moreover, the GMM priors were reparametrized by minimizing the negative log-likelihood with respect to an affine bijector and a Normal distribution:
\begin{equation}
    \hat{m} = \omega_{k} m + \phi_k
\end{equation}
Visualizations of reparametrizations for all GMM, as well as translation, theta and scale distributions are shown below:

\begin{figure}[htpb]
    \begin{subfigure}[T]{0.23\linewidth}
        \centering
        \includegraphics[width=\linewidth]{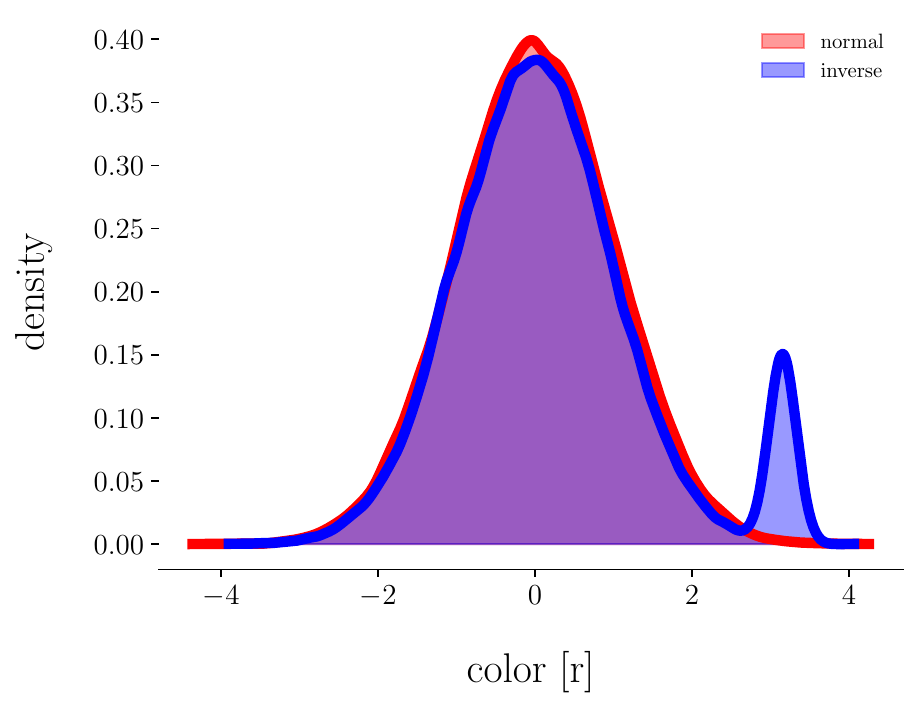}
        \caption{Reparametrization of the R prior}~\label{fig:reparametrization_R}
    \end{subfigure}
    \begin{subfigure}[T]{0.23\linewidth}
        \centering
        \includegraphics[width=\linewidth]{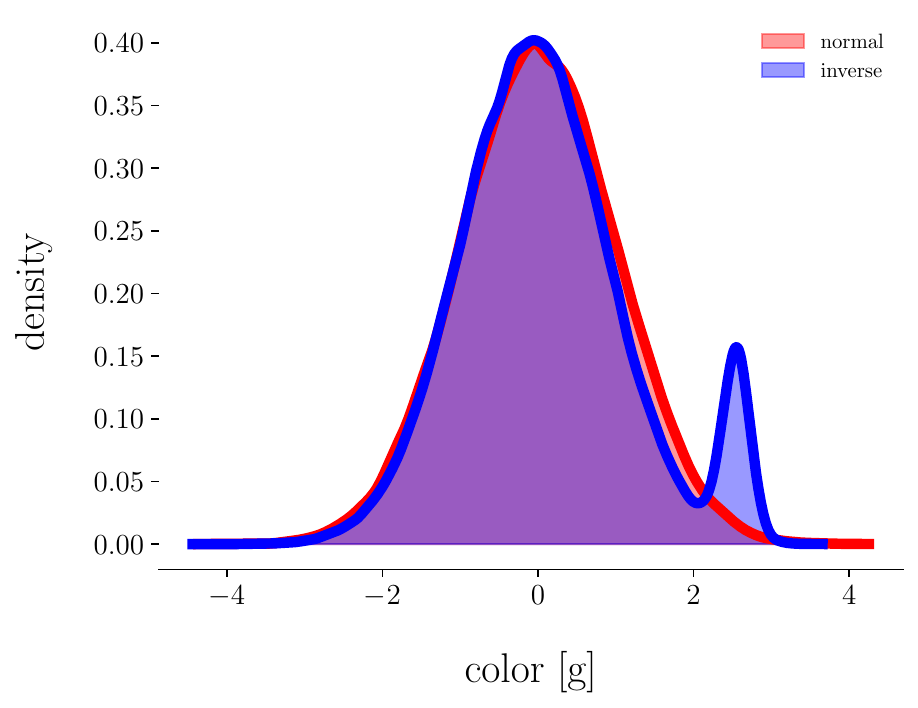}
        \caption{Reparametrization of the G prior}~\label{fig:reparametrization_G}
    \end{subfigure}
    \begin{subfigure}[T]{0.23\linewidth}
        \centering
        \includegraphics[width=\linewidth]{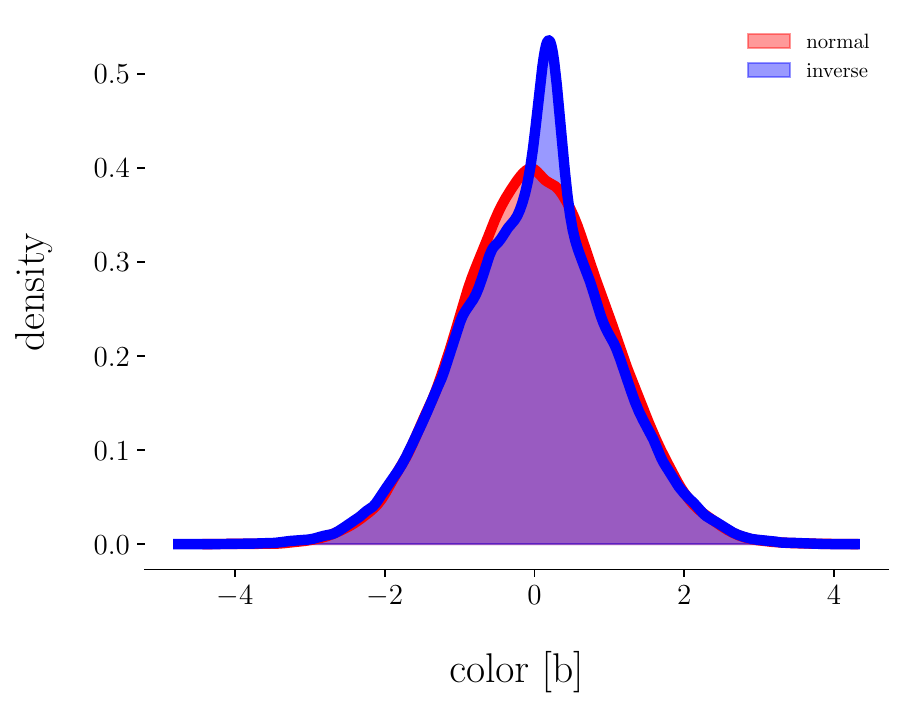}
        \caption{Reparametrization of the B prior}~\label{fig:reparametrization_B}
    \end{subfigure}
    \begin{subfigure}[T]{0.23\linewidth}
        \centering
        \includegraphics[width=\linewidth]{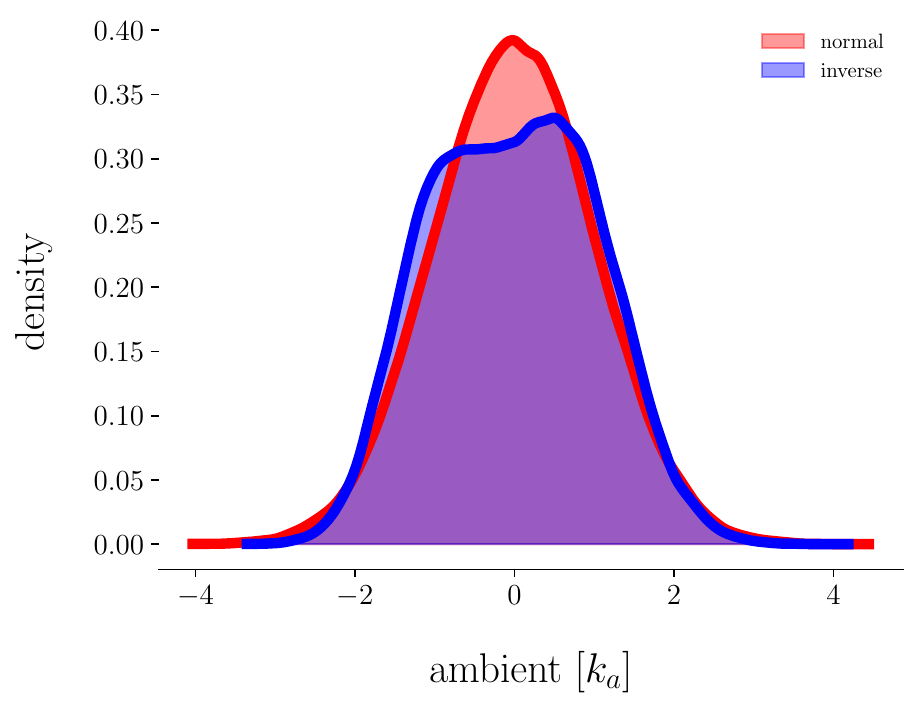}
        \caption{Reparametrization of the ambient prior}~\label{fig:reparametrization_ambient}
    \end{subfigure}
    \begin{subfigure}[T]{0.23\linewidth}
        \centering
        \includegraphics[width=\linewidth]{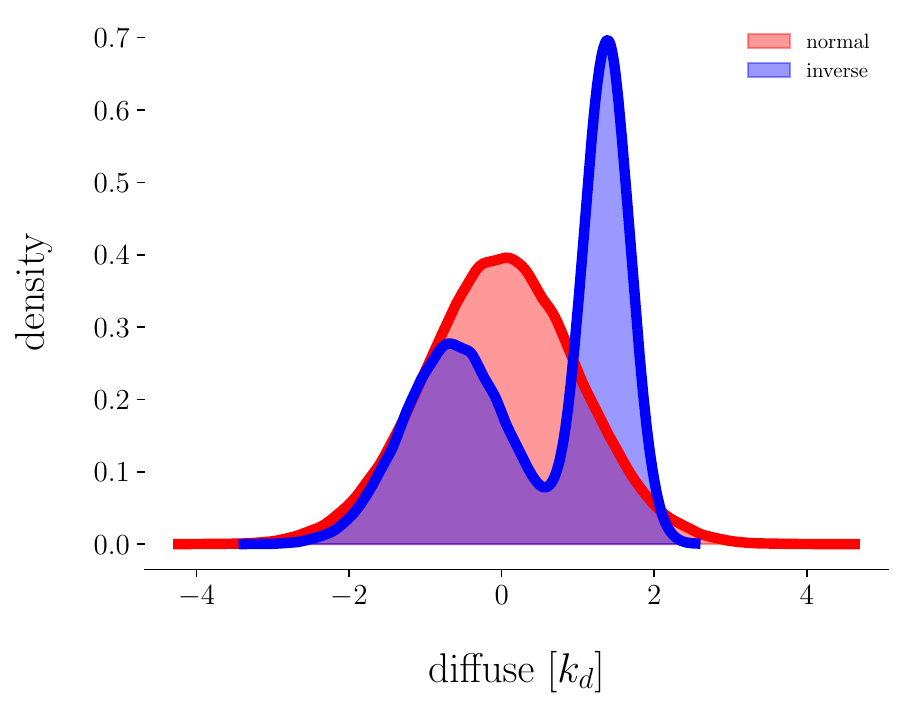}
        \caption{Reparametrization of the diffuse prior}~\label{fig:reparametrization_diffuse}
    \end{subfigure}
    \begin{subfigure}[T]{0.23\linewidth}
        \centering
        \includegraphics[width=\linewidth]{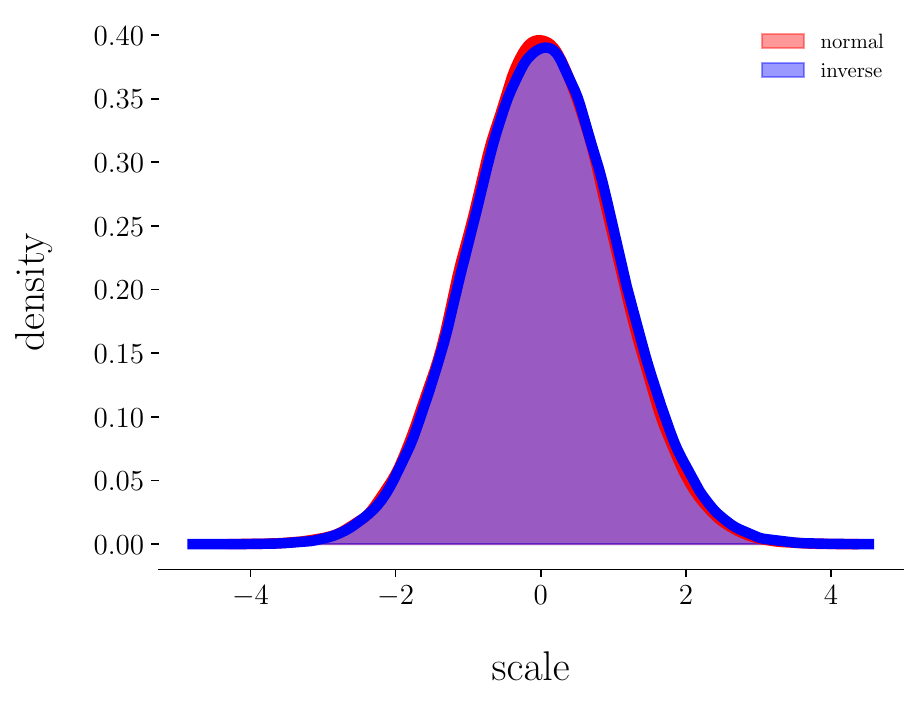}
        \caption{Reparametrization of the scale prior}~\label{fig:reparametrization_scale}
    \end{subfigure}
    \begin{subfigure}[T]{0.23\linewidth}
        \centering
        \includegraphics[width=\linewidth]{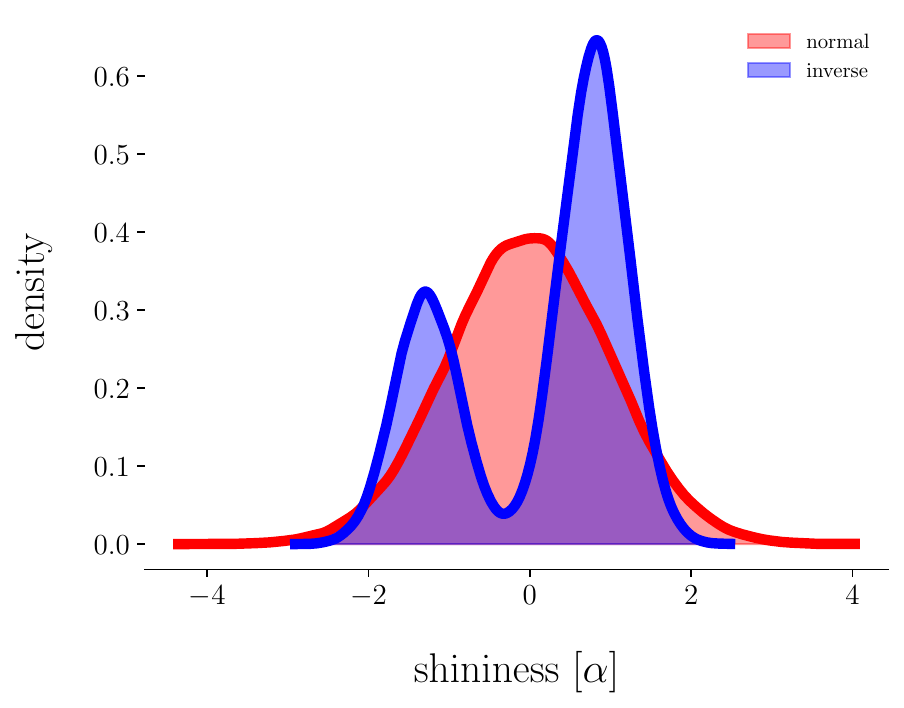}
        \caption{Reparametrization of the shininess prior}~\label{fig:reparametrization_shininess}
    \end{subfigure}
    \begin{subfigure}[T]{0.23\linewidth}
        \centering
        \includegraphics[width=\linewidth]{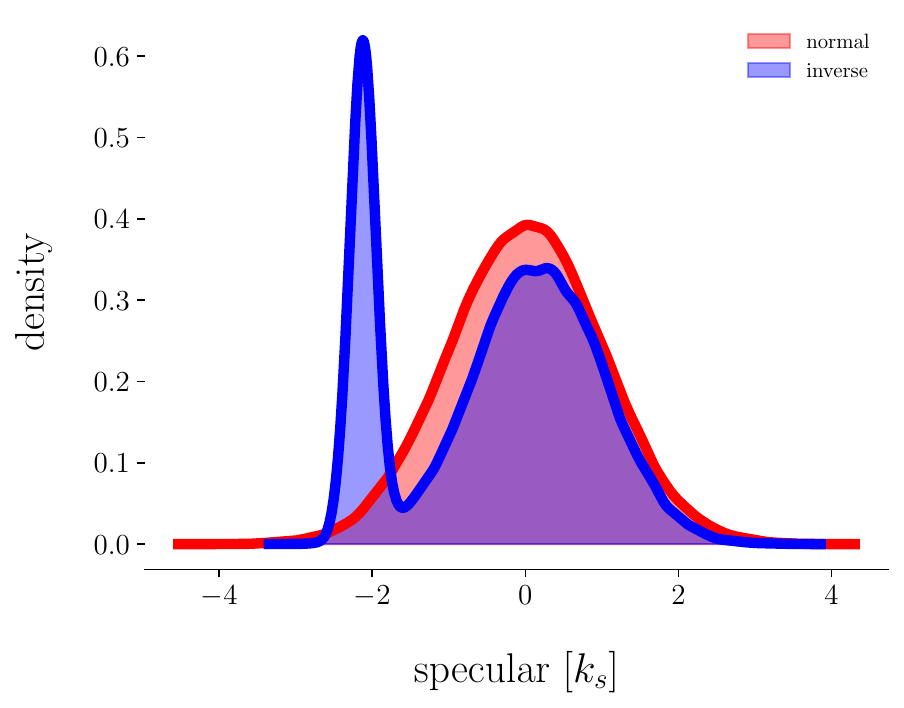}
        \caption{Reparametrization of the specular prior}~\label{fig:reparametrization_specular}
    \end{subfigure}
    \begin{subfigure}[T]{0.23\linewidth}
        \centering
        \includegraphics[width=\linewidth]{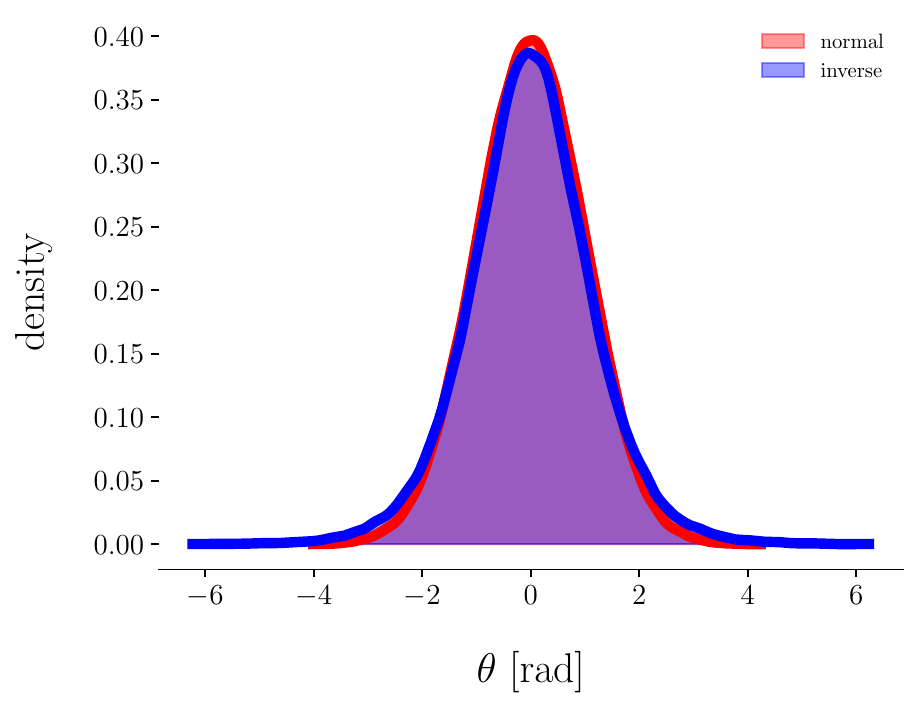}
        \caption{Reparametrization of the theta prior}~\label{fig:reparametrization_theta}
    \end{subfigure}
    \begin{subfigure}[T]{0.23\linewidth}
        \centering
        \includegraphics[width=\linewidth]{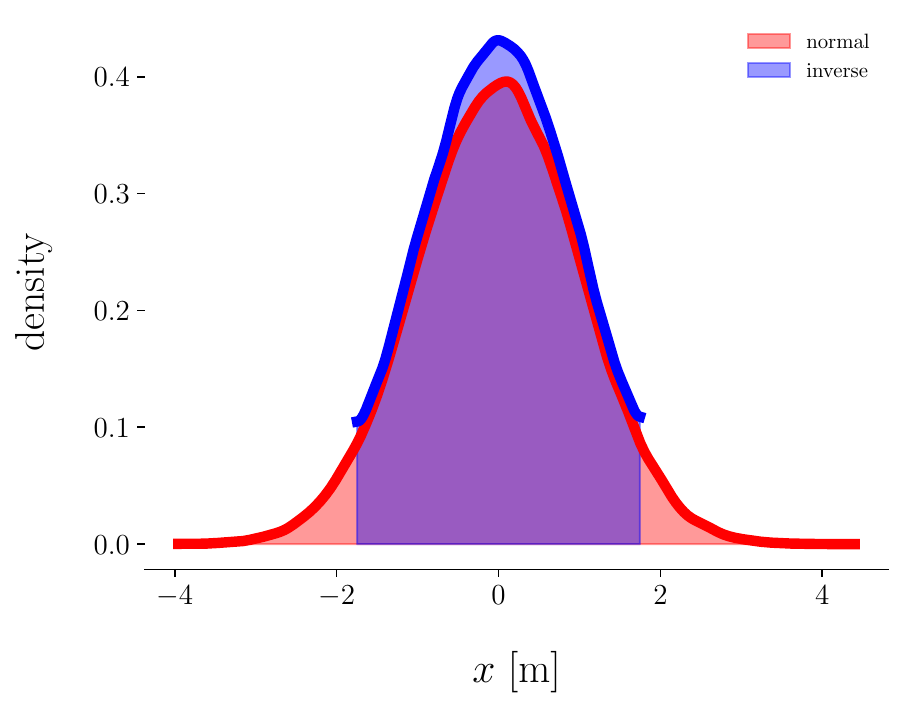}
        \caption{Reparametrization of the x translation prior}~\label{fig:reparametrization_x_translation}
    \end{subfigure}
    \begin{subfigure}[T]{0.23\linewidth}
        \centering
        \includegraphics[width=\linewidth]{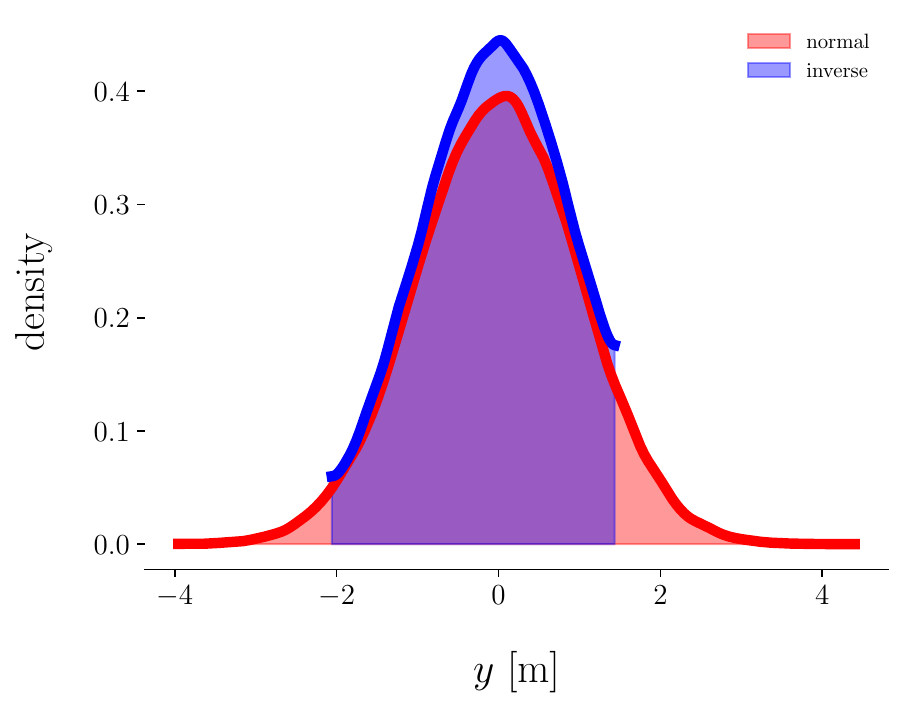}
        \caption{Reparametrization of the y translation prior}~\label{fig:reparametrization_y_translation}
    \end{subfigure}
    \caption{Reparametrized priors. Red indicates a Normal distribution. Blue the reparametrized prior.}
\end{figure}

\section{Posterior results}~\label{sec:posterior_results}
\begin{figure*}[htpb]
    \centering
    \includegraphics[width=0.95\textwidth]{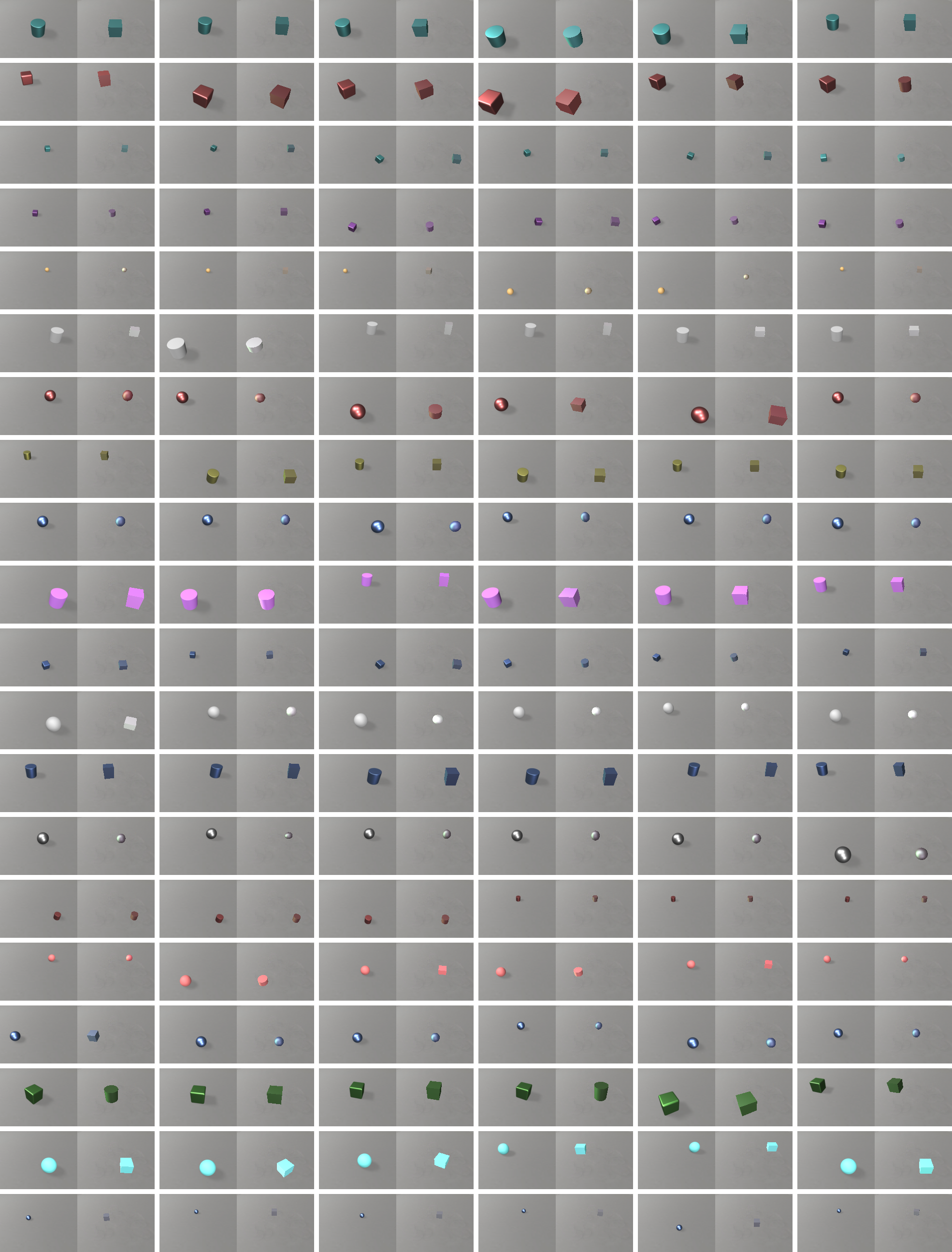}
    \caption{True image with NP3 median point estimate of the \fsclvr~test split}~\label{fig:median_point_estimates}
\end{figure*}

\begin{figure*}[htpb]
    \centering
    \includegraphics[width=0.95\textwidth]{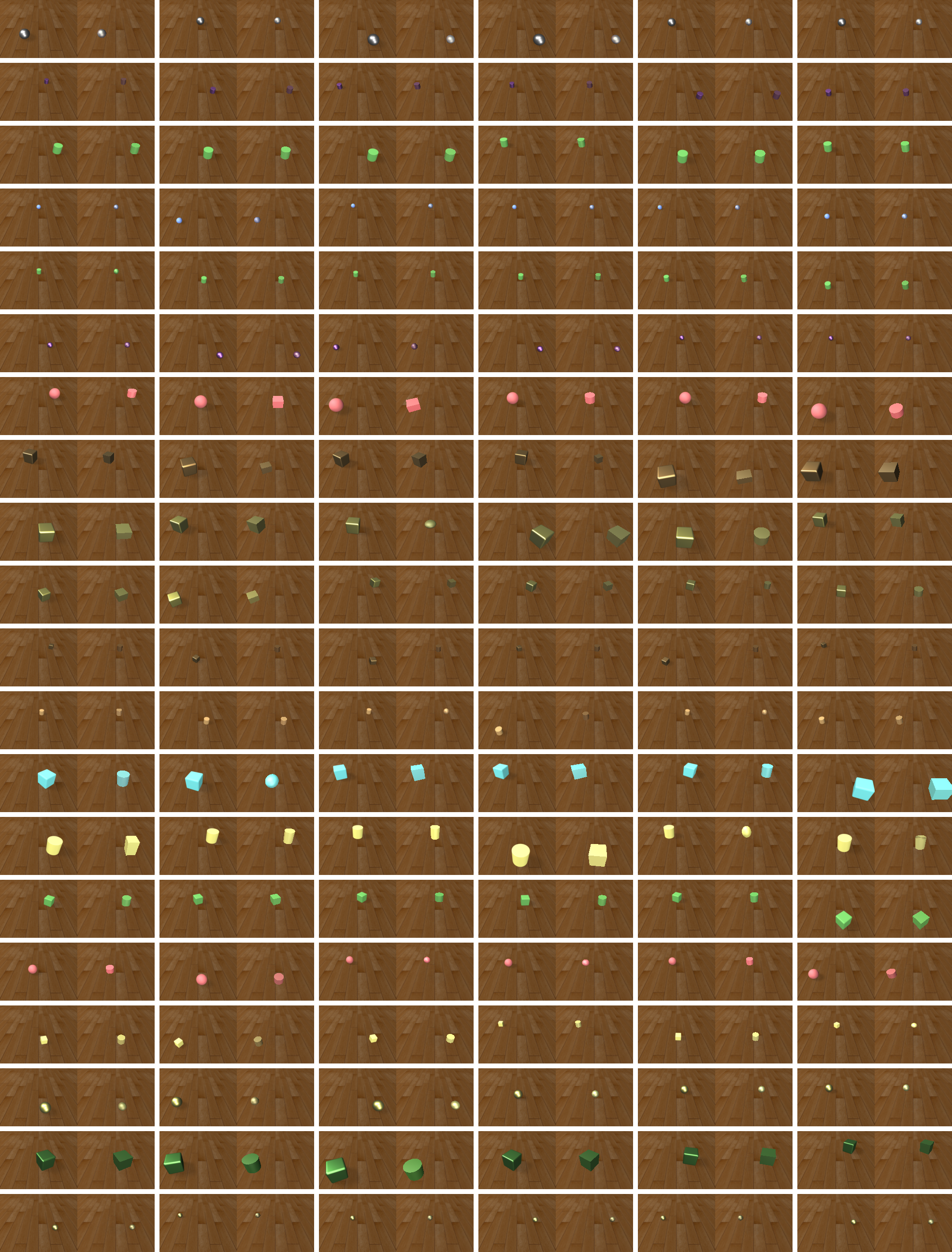}
    \caption{True image with NP3 median point estimate of the \fsclvrroom~test split}~\label{fig:median_point_estimates_room}
\end{figure*}

\begin{figure*}[htpb]
    \centering
    \includegraphics[width=0.95\textwidth]{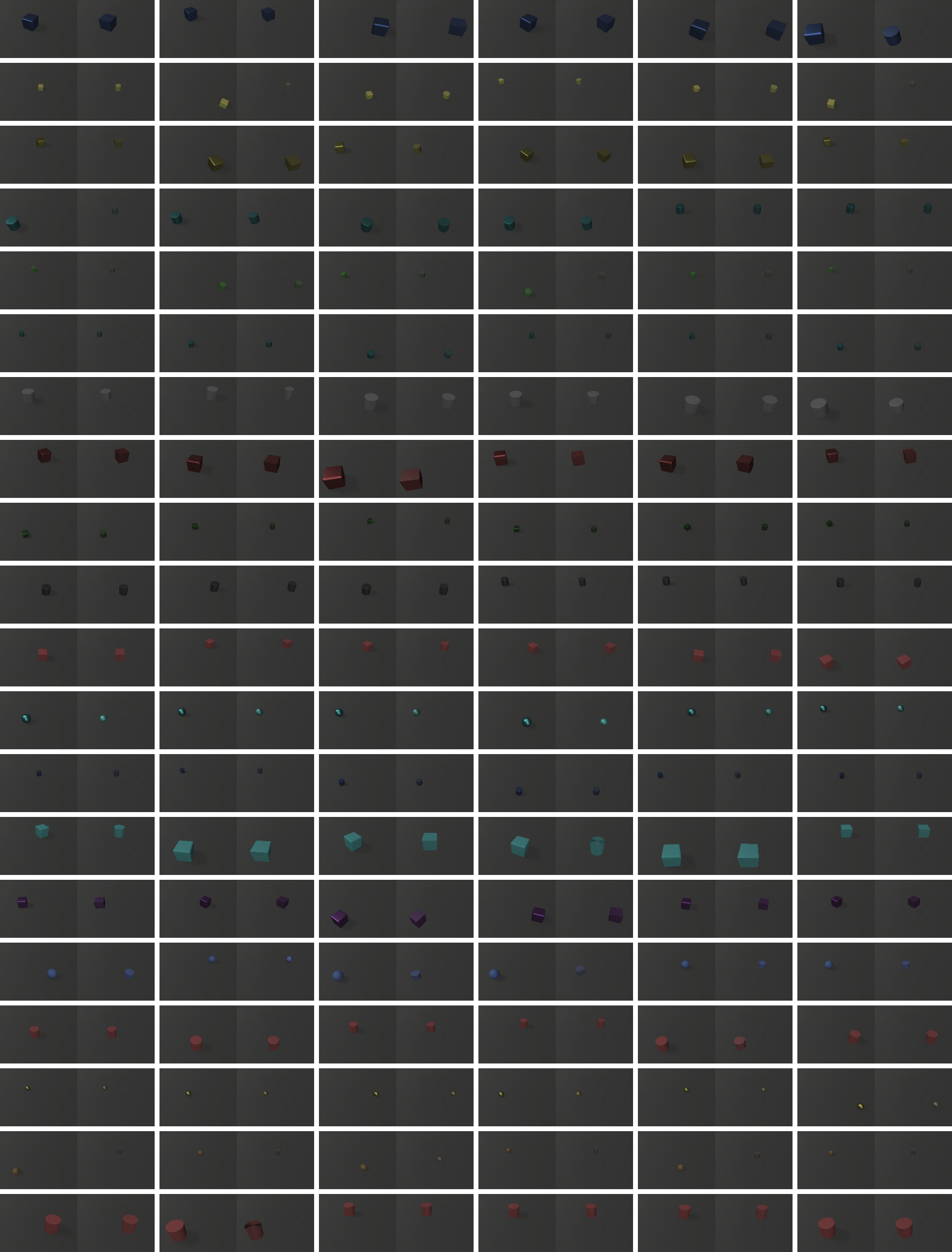}
    \caption{True image with NP3 median point estimate of the \fsclvrdark~test split}~\label{fig:median_point_estimates_dark}
\end{figure*}

\begin{figure*}[htpb]
    \centering
    \includegraphics[width=0.95\textwidth]{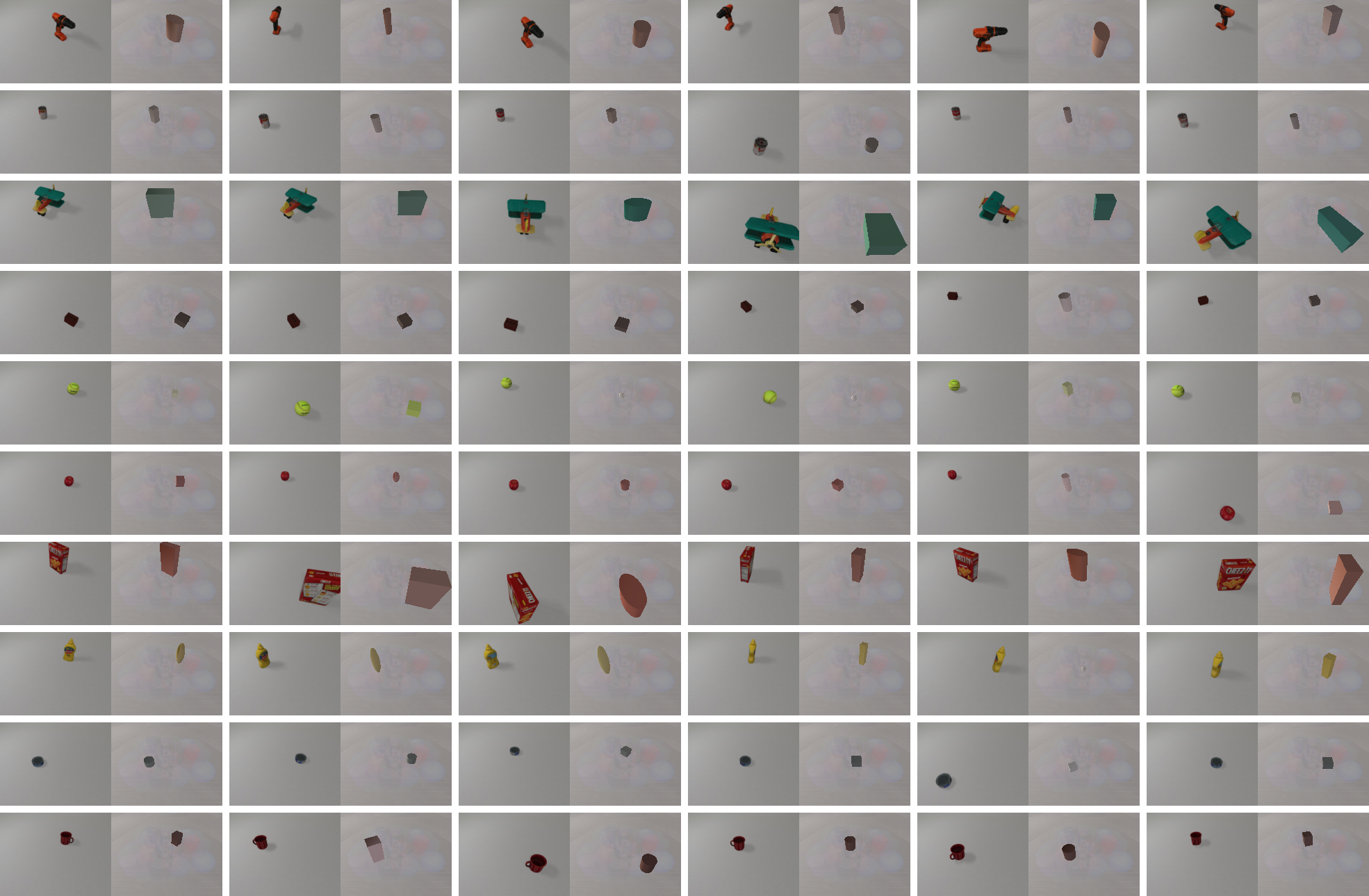}
    \caption{True image with NP3 median point estimate of the \ycbood~dataset}~\label{fig:median_point_estimates_ycbood}
\end{figure*}

    This section contains more posterior results.
Specifically, we show the median point estimates of all our predictions in Figure~\ref{fig:median_point_estimates} for the \fsclvr~dataset, in Figure~\ref{fig:median_point_estimates_room} for \fsclvrroom, Figure~\ref{fig:median_point_estimates_dark} for \fsclvrdark~and Figure~\ref{fig:median_point_estimates_ycbood} for the \ycbood~dataset.

\section{Prototypical program results}~\label{sec:supplementary_prototypical}
    In the image below we show our probabilistic prototypical program (P3) in graphical form.
This model is derived from our probabilistic generative model~\ref{fig:probabilistic_program} by removing the variables translation $\Shift$ and rotation $\Angle$ of our inverse graphics model.

\vspace{0.5cm}

\begin{tikzpicture}[scale=0.4, transform shape, node distance=0.4cm]
    \definecolor{conceptColor}{rgb}{0.90, 0.90, 0.90}

    \node[obs, draw=black!20, fill=black!20] (image) {$\mathcal{I}$};
    \node[det, draw=\mainColor!50, fill=\mainColor!51, above=0.22 of image] (render) {$\DiffGraphics$} ; %
    \node[latent, draw=conceptColor!80, fill=conceptColor!51, above=0.22 of render] (object) {$\mathcal{O}$};

    \node[latent, draw=conceptColor!75,fill=conceptColor!51, above=0.5 of object, xshift=-2.5cm] (material) {$\mathcal{M}$};
    \node[latent, draw=conceptColor!25,fill=conceptColor!51, above=0.5 of object, xshift=1.5cm]  (transform) {$\mathcal{A}$};
    \node[latent, draw=conceptColor!10, fill=conceptColor!51, above=of object, yshift=-0.5cm]  (shapearg) {$\boldsymbol{\kappa}$};

    \node[latent, draw=conceptColor!50,fill=conceptColor!51, above=0.5 of material, xshift=-2cm]  (color) {$c$};

    \node[latent, draw=conceptColor!35,fill=conceptColor!51, above=0.5 of material, xshift=-1cm]  (ambient) {$k^a$};    
    \node[latent, draw=conceptColor!60,fill=conceptColor!51, above=0.5 of material, xshift=0cm]  (specular) {$k^s$};   
    \node[latent, draw=conceptColor!25,fill=conceptColor!51, above=0.5 of material, xshift=1cm]  (diffuse) {$k^d$};     
    \node[latent, draw=conceptColor!50,fill=conceptColor!51, above=0.5 of material, xshift=2cm]  (shininess) {$\Shininess$}; 

    \node[latent, draw=conceptColor!50, fill=conceptColor!51, above=0.5 of transform, xshift=0cm]  (scale) {$\Scale$};

    \edge[-] {shapearg} {object} ;
    \edge[-] {shapearg, material, transform} {object} ;
    \edge[-] {ambient, diffuse, color, specular, shininess} {material} ;
    \edge[-] {scale} {transform} ;  
    \edge[-] {object} {render} ;
    \edge[-] {render} {image} ;

\end{tikzpicture}

\section{MAML and ProtoNets results}~\label{sec:sup_maml}
    We now present the additional experimental results of the CNN models used in MAML and ProtoNets.
This CNN model corresponds to the default architecture proposed in both papers.
Specifically it consists of 4 blocks, each having in sequence the following layers: \texttt{Conv2D} with 64 $3 \times 3$ filters, \texttt{BatchNorm}, \texttt{ReLU} and \texttt{MaxPool2D}.
Initially we did some preliminary experiments for determining the image resolutions.
We set the training iteration steps to 20K and the image sizes of the \fsclvr~dataset were downsampled to $28 \times 28$, $56 \times 56$, $84 \times 84$, $112 \times 112$, $168 \times 168$.
The corresponding test accuracies for these resolutions were respectively 67.9\%, 80.5\%, 71\%, 30.4\%, 22.4\% under the MAML training framework. 
From here it could be seen that the lower resolutions could achieve better performance. 
In all our experiments we trained the models for 60K iterations using the optimizer (ADAM) and learning rates (1e-3) suggested by the original authors.

\section{Evaluation of pose estimation}~\label{sec:sup_pose}
    In this section, we present our evaluation results of pose estimation with respect to the ADI metric (Average Distance of Indistinguishable Model Point)~\cite{hinterstoisser2013model} for our different version of models on our different test datasets  in Table~\ref{tab:pose_all}.
This ADI metric is commonly used for symmetric objects and measures the average deviation of the transformed model points to the closest model point.
The values in the Table~\ref{tab:pose_FS} and~\ref{tab:pose_YCBOOD} represents the mean value of the ADI error for our total 6 shots of each test class in our \fsclvr, \fsclvrroom, \fsclvrdark, and \ycbood~test dataset.

\begin{table}[!htb]
\scriptsize
\begin{tabular}{cc}
  \begin{subtable}{.69\linewidth}
  	\centering
    \begin{tabular}{l|c|c|c|c|c|c}
\hline
\multirow{2}{*}{\textbf{Test Classes}} & \multicolumn{2}{c|}{\underline{\textbf{\fsclvr}}} & \multicolumn{2}{c|}{\underline{\textbf{\fsclvrroom}}} & \multicolumn{2}{c}{\underline{\textbf{\fsclvrdark}}}\\
& \textbf{P3} & \textbf{NP3} & \textbf{P3} & \textbf{NP3} & \textbf{P3} & \textbf{NP3}\\
\hline
Class 0 & 0.0153 & 0.018  & 0.0075 & 0.0086 & 0.0235 & 0.0197 \\
Class 1 & 0.0174 & 0.0176 & 0.0369 & 0.0110 & 0.0791 & 0.0798 \\
Class 2 & 0.0092 & 0.0059 & 0.0072 & 0.0098 & 0.0694 & 0.0663 \\
Class 3 & 0.0136 & 0.0145 & 0.0054 & 0.0090 & 0.0781 & 0.0553 \\
Class 4 & 0.0697 & 0.0391 & 0.0072 & 0.0069 & 0.086  & 0.086  \\
Class 5 & 0.0258 & 0.0265 & 0.0215 & 0.0200 & 0.0852 & 0.0856 \\
Class 6 & 0.0045 & 0.0058 & 0.0109 & 0.0115 & 0.0509 & 0.0497 \\
Class 7 & 0.011  & 0.0108 & 0.0154 & 0.0181 & 0.0148 & 0.0126 \\
Class 8 & 0.0031 & 0.0054 & 0.0203 & 0.0228 & 0.042  & 0.0407 \\
Class 9 & 0.0119 & 0.0111 & 0.0211 & 0.0168 & 0.0394 & 0.0364 \\
Class 10 & 0.0088 & 0.0104 & 0.0729 & 0.0725 & 0.0515 & 0.0184 \\
Class 11 & 0.0115 & 0.0117 & 0.0246 & 0.0225 & 0.0642 & 0.0263 \\
Class 12 & 0.0134 & 0.0125 & 0.0169 & 0.0205 & 0.075  & 0.0745 \\
Class 13 & 0.0059 & 0.0057 & 0.018  & 0.0197 & 0.0168 & 0.0146 \\
Class 14 & 0.007  & 0.0067 & 0.0139 & 0.0111 & 0.055  & 0.0454 \\
Class 15 & 0.0095 & 0.0088 & 0.0111 & 0.0140 & 0.0697 & 0.053  \\
Class 16 & 0.0076 & 0.0098 & 0.019  & 0.0200 & 0.0552 & 0.0148 \\
Class 17 & 0.0139 & 0.0119 & 0.013  & 0.0177 & 0.0849 & 0.0857 \\
Class 18 & 0.0106 & 0.0095 & 0.0173 & 0.0185 & 0.0772 & 0.0779 \\
Class 19 & 0.0344 & 0.024  & 0.0095 & 0.0080 & 0.024  & 0.0175 \\
\hline
\end{tabular}
    \caption{Evaluation on the \fsclvr , \fsclvrroom , and \fsclvrdark~ test dataset}~\label{tab:pose_FS}
  \end{subtable}
  &
  \begin{subtable}{.3\linewidth}
  \centering
    \begin{tabular}{l|c|c}
\hline
\multirow{2}{*}{\textbf{Test Classes}} & \multicolumn{2}{c}{\underline{\textbf{\ycbood}}}\\
& \textbf{P3} & \textbf{NP3} \\
\hline
power-drill & 0.0274 & 0.0272 \\
tomato-soup & 0.0219 & 0.0218 \\
airplane-A & 0.0212 & 0.0198 \\
foam-brick & 0.0259 & 0.0274 \\
softball & 0.0768 & 0.0749 \\
apple & 0.0466 & 0.0435 \\
cracker-box & 0.0313 & 0.0326 \\
mustard-bottle & 0.0534 & 0.0494 \\
tuna-fish-can & 0.0312 & 0.0333 \\
mug & 0.0242 & 0.0231 \\
\hline
\end{tabular}
    \caption{Evaluation on the \ycbood~test dataset}~\label{tab:pose_YCBOOD}
  \end{subtable}
\end{tabular}
\caption{Evaluation of pose estimation with ADI metric on our datasets}
~\label{tab:pose_all}
\end{table}

\end{document}